\documentclass[sigconf,acmthm=false]{acmart}

\setcopyright{none}
\renewcommand\footnotetextcopyrightpermission[1]{}
\acmConference[HSCC+ICCPS '27]{30th ACM/IEEE International Conference on Cyber-Physical and Autonomous Systems}{May 17--20, 2027}{Boulder, CO, USA}
\acmYear{2027}

\usepackage{subcaption}
\usepackage{siunitx}
\usepackage{multirow}
\usepackage{rotating}
\let\acmproof\proof
\let\endacmproof\endproof
\let\proof\relax 
\let\endproof\relax
\usepackage{amsthm}
\usepackage{amsfonts,bm,mathtools}
\usepackage[ruled,linesnumbered, noend]{algorithm2e}

\newtheorem{theorem}{Theorem}%[section]
\newtheorem{proposition}[theorem]{Proposition}%

\newtheorem{lemma}[theorem]{Lemma}

\newtheorem{definition}{Definition}%[theorem]
\newtheorem{assumption}{Assumption}%[theorem]
\newtheorem{remark}{Remark}%[theorem]
\newtheorem{problem}{Problem} %[section]
\def\tanh{{\mathrm{Tanh}}}

\def\Prob{\mathbb{P}}
\def\ProbQ{\mathbb{Q}}
\def\ProbG{\mathbb{G}}

\def\ProbHat{\widehat{\mathbb{P}}}

\def\expect{{\mathbb{E}}}

\def\wasserstein{{\mathbb{W}}}

\def\indicator{{\mathbf{1}}}

\def\ambesc{{\mathcal{A}}}

\newcommand{\norm}[1]{\left\lVert#1\right\rVert}

\def\realNum{{\mathbb{R}}}
\def\natNum{\mathbb{N}}

\def\ball{{\mathbb{B}}}

\def\sB{{\mathcal{B}}}

\def\sD{{\mathcal{D}}}
\def\sE{{\mathcal{E}}}
\def\sF{{\mathcal{F}}}

\def\sL{{\mathcal{L}}}

\def\sO{{\mathcal{O}}}
\def\sP{{\mathcal{P}}}

\def\sS{{\mathcal{S}}}
\def\sT{{\mathcal{T}}}
\def\sU{{\mathcal{U}}}

\def\sW{{\mathcal{W}}}
\def\sX{{\mathcal{X}}}
\def\sY{{\mathcal{Y}}}

\def\vpi{{{\pi}}}

\def\vomega{{{\omega}}}

\def\vc{{{c}}}

\def\vu{{{u}}}

\def\vx{{{x}}}

\newcommand{\evpi}[1]{{\vpi^{(#1)}}}

\newcommand{\evomega}[1]{{\vomega^{(#1)}}}

\def\lowerb{p_{\ell}}
\newcommand{\evlowerb}[1]{{\lowerb^{(#1)}}}
\def\upperb{p_{u}}
\newcommand{\evupperb}[1]{{\upperb^{(#1)}}}

\def\samplesVector{\mathcal{D}_N}
\newcommand{\evgamma}[1]{{\gamma^{(#1)}}}

\newcommand{\citehere}{\textcolor{red}{\textbf{citehere}}}
\newcommand{\refhere}{\textcolor{red}{\textbf{refhere}}}
\let\proof\acmproof
\let\endproof\endacmproof

\newcommand{\LL}[1]{\textcolor{red}{[LL: #1]}}

\newcommand{\EF}[1]{\textcolor{red}{[EF: #1]}}

\begin{document}

\title{Learn-To-Propagate: Distributionally Robust Motion Planning for Nonlinear Uncertain Systems}

% Hidden by the 'anonymous' option. Fill in before the camera-ready version.
\author{Eduardo Figueiredo}
\affiliation{%
  \institution{Delft University of Technology}
  \city{Delft}
  \country{Netherlands}}
\email{e.figueiredo@tudelft.nl}

\author{Adrien Banse}
\affiliation{%
  \institution{ETH Zurich}
  \city{Zurich}
  \country{Switzerland}}
\email{abanse@ethz.ch}

\author{Qi Heng Ho}
\affiliation{%
  \institution{Virginia Tech}
  \city{Virginia}
  \country{USA}}
\email{qihengho@vt.edu}

\author{Raphaël Jungers}
\affiliation{%
  \institution{UCLouvain}
  \city{Louvain-la-Neuve}
  \country{Belgium}}
\email{raphael.jungers@uclouvain.be}

\author{Morteza Lahijanian}
\affiliation{%
  \institution{University of Colorado Boulder}
  \city{Boulder}
  \country{USA}}
\email{morteza.lahijanian@colorado.edu}

\author{Luca Laurenti}
\affiliation{%
  \institution{Delft University of Technology}
  \city{Delft}
  \country{Netherlands}}
\email{l.laurenti@tudelft.nl}

\begin{abstract}
    Motion planning for stochastic systems aims to find a sequence of control inputs that drives a system to a goal set while avoiding unsafe states with high probability. Existing approaches, however, are either restricted to linear dynamics with perfect knowledge of the underlying distributions or lack formal correctness guarantees. 
To address these limitations, we propose \emph{Learn-To-Propagate Distributionally-Robust Forest (L2P-DR-Forest)}, a distributionally-robust sampling-based planning algorithm for \emph{nonlinear} stochastic systems with \emph{unknown} initial state and process noise distributions. Our approach leverages recent advances in optimal transport to learn ambiguity sets that contain the true system distributions with high confidence. We develop an efficient method for propagating these sets through the system dynamics and for verifying their safety. We then integrate these techniques into a novel sampling-based planning algorithm to grow a forest of search trees in the system's state space while accounting for and mitigating the growth of ambiguity-sets radii. We prove that the algorithm is correct, i.e., every returned plan satisfies the specifications with the prescribed confidence, and, under some feasibility and distributional assumptions, is probabilistically complete.
Evaluations on several robotic benchmarks demonstrate that our algorithm significantly outperforms existing methods
while providing formal guarantees with high confidence.
\end{abstract}

% ACM CCS concepts: check them, and generate the matching CCSXML block, at https://dl.acm.org/ccs


\maketitle

%===============================================================================

\section{Introduction}

Motion planning has become a central problem in robotics, enabling autonomous systems to operate safely in complex environments while avoiding obstacles. Applications span domains such as autonomous driving, aerial navigation, and robotic surgery. In this context, sampling-based algorithms have emerged as particularly successful approaches \cite{cortes2020samplingbased}, where samples from the state and control spaces are used to efficiently construct search trees and compute collision-free trajectories. However, real-world robotic systems are often subject to uncertainty arising from process and measurement noise, external disturbances, and imperfect knowledge of the system dynamics. Unfortunately, existing approaches either rely on linear-Gaussian assumptions to obtain tractable safety guarantees or accommodate more general uncertainty at the expense of formal correctness guarantees \cite{blackmore2011chance, luders2010chance, luders2013robust, ho2022gaussian, prattis2026stochastic, dai2019chance, wang2020non}.
This work aims to bridge this gap by developing a motion planning framework for uncertain systems that provides both correctness and completeness guarantees.

Among approaches that seek formal safety guarantees under uncertain disturbances, two broad directions have emerged. The first assumes that disturbances have bounded support, reducing safety checks to a worst-case reachability problem \cite{wu2022robust, danielson2020robust}. While effective in some settings, these methods often become overly conservative when the support of the uncertainty is large. A second direction relies on distributionally robust optimization (DRO) \cite{mohajerin2018data}, constructing sets of plausible probability distributions, referred to as ambiguity sets, and evaluating worst-case probabilities over these sets \cite{summers2018distributionally, hakobyan2020wasserstein, renganathan2020towards, lindemann2021robust, gracia2026provablySafe}. Yet, most existing DRO-based approaches are limited to linear or locally linearized systems, can be overly conservative, and either require a predefined ambiguity set or face challenges in learning sufficiently tight ambiguity sets from data. To address these limitations, we show how recent advances in Wasserstein-based distribution learning \cite{figueiredo2026efficient, fournier2015rate} and nonlinear uncertainty propagation \cite{figueiredo2025efficient} can be combined to enable safe motion planning for nonlinear stochastic systems with unknown disturbance distributions.

In this paper, we propose \emph{Learn-To-Propagate Distributionally-Robust Forest (L2P-DR-Forest)}, a novel and computationally efficient motion-planning algorithm that combines DRO, sampling-based planning, and nonlinear uncertainty propagation to handle stochastic systems with unknown initial state and disturbance distributions. Leveraging recent results in optimal transport, our approach estimates a compact ambiguity set centered around a discrete distribution that is guaranteed, with high probability, to contain the true underlying system distributions. 
Critically, Wasserstein ambiguity sets admit a highly tractable propagation rule through Lipschitz continuous dynamics, with their radii expanding at most linearly with the Lipschitz constant of the dynamics. Consequently, these sets can be efficiently propagated while preserving computable bounds on distributional uncertainty, avoiding the need for relearning at every propagation step. This property enables rapid exploration of the planning tree and substantially reduces the computational burden associated with exhaustive sampling. Combined with existing DRO results \cite{gao2023distributionally}, this allows us to synthesize motion plans that carry formal collision-avoidance guarantees despite underlying distributional uncertainty. 

L2P-DR-Forest grows a forest of trees in stages. Each new tree is rooted at a node of an existing tree, and the ambiguity set at this root is re-learned from samples rolled out through the controller leading to that node. Thus, the planner propagates ambiguity sets cheaply within each tree and re-learns tight ones only at the selected restart nodes.
We prove that the algorithm is \emph{correct}, i.e., any controller it returns solves the planning problem with the prescribed confidence, and \emph{probabilistically complete}, i.e., if a control sequence is robustly feasible with some margin and the noise is sub-Gaussian, then, run for an unbounded number of stages with growing sample sizes, it returns a controller with probability one.
We evaluate L2P-DR-Forest on a range of nonlinear robotics benchmarks. Our experiments demonstrate that the proposed approach achieves high task-success rates while enforcing safe controllers. In contrast, comparable methods that do not provide formal safety guarantees exhibit substantially lower reliability.

In summary, our main contributions are:
\begin{itemize}
  \item A novel distributionally robust sampling-based motion planning algorithm, \emph{L2P-DR-Forest}, which, to the best of our knowledge, is the first motion planner that provides formal, high-confidence safety guarantees for nonlinear systems with unknown distributions. 
    \item An extension of data-driven distributionally robust algorithms to learn distributions with unbounded support from data, which is critical to formally reduce the uncertainty in our motion planner over forest expansion.
    \item A formal proof that L2P-DR-Forest is \emph{correct} and \emph{probabilistically complete}.
    \item An empirical evaluation including a neural model of a tractor trailer showing that our approach substantially outperforms comparable methods both in terms of success rate and planning time.
\end{itemize}

\paragraph{Related Works}
Motion planning under uncertainty is commonly posed as a chance-constrained problem \cite{blackmore2011chance, luders2010chance, luders2013robust, kothari2013probabilistically, dai2019chance}, where the probability of unsafety (e.g., collision) must satisfy a given upper bound at every time step.
There are two inherent challenges in such problems: (i) uncertainty propagation through the dynamics, and (ii) bounding the probability of unsafety.
Most of the existing approaches assume linear or linearized dynamics and Gaussian uncertainty to obtain tractable probability bounds or approximations. Extensions to non-Gaussian uncertainty use moment-based bounds \cite{wang2020non, jasour2019risk}, which provide formal guarantees under restrictive assumptions, such as access to the relevant moments and their dynamics. Even when these assumptions hold, such methods can be overly conservative or computationally expensive.

For nonlinear systems, a common alternative is to assess safety statistically from sampled rollouts. This accommodates general dynamics and noise distributions, but naive Monte Carlo estimation can require many samples to resolve small collision probabilities, motivating variance-reduction techniques \cite{janson2017monte}. Sampling also underpins model predictive control methods such as MPPI \cite{williams2017model}, whose original formulation does not provide finite-sample probabilistic safety certificates. Among sampling-based planners, $\epsilon$-RandUP-RRT \cite{wu2022robust, lew2022randup}, which we use as a baseline in Section~\ref{section:experimental-results}, checks collision avoidance using $\epsilon$-padded convex hulls of propagated particles. The finite-sample guarantees of its underlying reachable-set estimator depend on bounds on quantities such as Lipschitz constants, packing numbers, and sampling coverage, which can be difficult to obtain tightly in practice. 

To mitigate these issues, scenario-based planning \cite{de2025scenario} has been introduced.  It provides distribution-free, finite-sample guarantees from i.i.d. samples drawn from the true uncertainty distribution.  These methods have been applied to nonlinear model predictive control and motion planning among uncertain obstacles, but they assume the robot dynamics are deterministic. These guarantees extend to nonconvex decision problems, although certification requires suitable feasibility conditions and bounds on the number of supporting scenarios. In contrast, in this work, we focus on offline, sampling-based planning for nonlinear systems with unknown disturbance distributions, with probabilistic safety guarantees derived from finite data.

Distributionally robust optimization addresses uncertainty about the underlying probability law by optimizing against the worst-case distribution within a set of plausible distributions, called an ambiguity set. In motion planning, ambiguity sets have been defined through means and covariances for linear systems with additive process noise \cite{summers2018distributionally}, and as Wasserstein balls centered on empirical distributions constructed from samples of obstacle motion within model predictive control \cite{hakobyan2020wasserstein}. Closely related work \cite{gracia2026provablySafe} learns Wasserstein ambiguity tubes from trajectory data under unknown disturbance distributions and incorporates them into a probabilistically complete sampling-based planner. Its construction exploits linear or feedback-linearizable dynamics to decouple uncertainty propagation from nominal planning, allowing the ambiguity tube to be learned offline and reused during search. In contrast, we propagate learned ambiguity sets through nonlinear Lipschitz continuous dynamics and re-learn them during search, using a forest of trees to mitigate growth in their radii.

\section{Preliminaries}\label{section:preliminaries}
\paragraph{Notation.} For a vector $x \in \realNum^d$,  we denote by $x^{(i)}$ its $i$-th element. For a set $\sX \subseteq \realNum^d$, the indicator function for $\sX$  is denoted as $\indicator_{\sX}(x) \coloneqq 1 \text{ if } x \in \sX \text{; otherwise } 0$. We represent the set of indexes $\{1, \dots, M\}$ by $[M]$.
Given a Borel measurable set $\sX \subseteq \realNum^d$, we denote by $\sB(\sX)$ the Borel sigma algebra over $\sX$ and by $\sP(\sX)$ the set of probability distributions on $\sX$. The symbol $\otimes$ is used to represent the product measure between two distributions. The restriction of a distribution $\Prob\in \sP(\sX)$ to the subspace $\sX_c\subseteq \sX$ is denoted by $\Prob|_{\sX_c}$ and defined as $\Prob|_{\sX_c}(A)=\Prob(A \cap \sX_c) / \Prob(\sX_c)$ for all $A \in \sB(\sX)$.
For a probability distribution ${\Prob} \in \sP(\sX)$ and a measurable function $g: \sX \rightarrow \sY \subseteq \realNum^{q}$, we denote the push-forward measure of ${\Prob}$ by $g$ as $g \# {\Prob}$ such that for all $A \in\sB(\sY)$, $(g \# \Prob)(A) \coloneqq \Prob(g^{-1}(A))$. We note that $g \# \Prob$ is still a probability distribution such that $g \# \Prob \in \sP(\sY)$.

\paragraph{Wasserstein distance.} We start by recalling the definition of the \emph{Wasserstein distance} and \emph{Wasserstein ambiguity set}, which we use to model distributional uncertainty in our framework. 
Let  $\rho \geq 1$, $\mathcal{X}\subseteq \mathbb{R}^d$, and define $\sP_{\rho}(\sX)$ as the set of probability distributions with finite $\rho$-th moment, i.e., all $\Prob \in \sP(\sX)$ such that $\int_{\sX} \norm{x}^\rho d\Prob(x) < \infty$, where $\norm{\cdot}$ is any fixed norm on $\realNum^d$.

\begin{definition}[$\rho$-Wasserstein distance]\label{def:wasserstein-distance}
For distributions $\mathbb{P}, \mathbb{Q} \in \sP_{\rho}(\sX)$, the $\rho$-Wasserstein distance between $\mathbb{P}$ and $\mathbb{Q}$ is defined as
\begin{equation} 
    \wasserstein_{\rho}(\Prob, \ProbQ) \coloneqq \left( \inf_{\gamma \in \Gamma(\mathbb{P}, \mathbb{Q})} \int_{\sX \times \sX} \norm{x-y}^{\rho} d\gamma(x, y) \right)^{\frac{1}{\rho}}
\end{equation}
where $\Gamma(\mathbb{P}, \mathbb{Q}) \subset \sP_{\rho}(\sX \times \sX)$ represents the set of joint probability distributions with given marginals $\mathbb{P}$ and $\mathbb{Q}$ (also known as \emph{couplings} between $\mathbb{P}$ and $\mathbb{Q}$), i.e., for all $\gamma \in \Gamma(\mathbb{P}, \mathbb{Q})$, it holds that
$
\gamma(A \times \sX) = \mathbb{P}(A), \; \gamma(\sX \times A) = \mathbb{Q}(A)
$
for all sets $A \in \sB(\sX)$.
\end{definition}
Additionally, we define the $\rho$-Wasserstein ambiguity set (ball) of radius $\epsilon \geq 0$ centered at the probability distribution $\Prob \in \sP_\rho(\sX)$ as:
\begin{align}
    \mathbb{B}_\epsilon(\Prob) \coloneqq \Big\{ \ProbQ \in \sP_\rho(\sX) : \wasserstein_\rho(\Prob, \ProbQ) \leq \epsilon \Big\}.
\end{align}
That is, $ \mathbb{B}_\epsilon(\Prob)$ contains all the probability distributions within $\epsilon$ distance of $\Prob$ according to $\wasserstein_\rho$.

\section{Problem Formulation}\label{section:problem-statement}

We consider a robot with motion uncertainty, whose state evolution is described by the stochastic process
\begin{equation}\label{eq:system_dynamics}
    \vx_{k+1} = f(\vx_k, \vu_k, \vomega_k), \quad \vx_0\sim\Prob_{0}, \quad \vomega_k\sim\Prob_{\vomega},
\end{equation}
where $x_k \in \sX \subset \realNum^d$ is the state, $u_k \in \sU \subset \realNum^q$
is the control action, and $\vomega_k \in \sW \subseteq \realNum^w$ is the process noise at time step $k$. Both $\sX$ and $\sU$ are assumed to be bounded sets.
Vector field \(f: \sX\times \sU \times \sW \to \sX\)
is a (possibly nonlinear) function, jointly Lipschitz continuous w.r.t. $(x,u,\vomega)$.
% , representing the one-step dynamics of System~\eqref{eq:system_dynamics}. 
The initial state $\vx_0$ of the robot is uncertain and distributed according to $\Prob_0$. 
In this paper, we make the following assumption about $\Prob_0$ and noise distribution $\Prob_\omega$.
\begin{assumption}\label{assumption:main-assumptions}
    We assume that, for any $k\geq 0$, $\vomega_k \sim \Prob_{\vomega}$ are i.i.d. and independent from $x_0 \sim \Prob_{0}$, with  $\Prob_{0}\in \sP(\sX)$ and $\Prob_{\vomega}\in \sP(\sW)$ being \emph{unknown}, but with finite $\rho$-moments. Furthermore, we assume access to two fixed, offline datasets, mutually independent of each other: $N_{x_0}>0$ i.i.d.\ samples $\sD_{x_0}$ of $x_0 \sim \Prob_0$, and $N_\vomega>0$ i.i.d.\ samples $\sD_\vomega$ of $\vomega \sim \Prob_\vomega$.
\end{assumption}
The finite $\rho$-moment assumption ensures that the $\rho$-Wasserstein distance between the probability measures considered in this work is finite, while the i.i.d. assumption excludes temporal correlations in the process noise. The availability of samples from $\Prob_{\vomega}$, although common in data-driven stochastic and distributionally robust settings
\cite{badings2023robust,gracia2024data,gracia2026provablySafe}, is stronger. Nevertheless, it
can be practically satisfied through, e.g., observations of the state and control. In practice, it suffices for $f$ to be injective over only a subset of $\sX$ as discussed in~\cite{gracia2024data}.
% 
% In this paper, it is required to enable the construction of data-driven Wasserstein ambiguity sets that contain the unknown true distributions with high confidence.

To make probabilistic statements about System~\eqref{eq:system_dynamics}, we define the probability law induced by a fixed control sequence. In particular, for $k>1$ and a given sequence of control inputs
$\bm{u}_{k-1}:=(u_0,\dots,u_{k-1})\in \mathcal U^k, $ the distribution of System \eqref{eq:system_dynamics} at time $k$ is denoted as \(\Prob^{\bm u_{k-1}}_{x_k}\) and is defined by propagating the noise and input distribution through the dynamics $k$ times using $\bm{u}_{k-1}$, i.e., 
\begin{align*}
\Prob^{\bm u_0}_{x_1}
&= f(\cdot,u_0,\cdot)\#(\Prob_0\otimes \Prob_\omega), \\
\Prob^{\bm u_{k-1}}_{x_k}
&= f(\cdot,u_{k-1},\cdot)\#\left(
\Prob^{\bm u_{k-2}}_{x_{k-1}}\otimes \Prob_\omega
\right) \quad \text{ for } k > 1.
\end{align*}
%Equivalently, for any measurable set \(\sA\subseteq\sX\),
%\begin{equation}\label{eq:recursive-def-state-distribution}
%\Prob^{\bm u_{k-1}}_{x_k}(\sA)
%=
%\int_{\sX}\int_{\sW}
%\bm\indicator_{\sA}
%\big(f(x_{k-1},u_{k-1},\omega_{k-1})\big)
%\Prob_\omega(\mathrm d\omega_{k-1})
%\Prob^{\bm u_{k-2}}_{x_{k-1}}(\mathrm d x_{k-%1}).
%\end{equation}

The robot operates in a workspace with obstacles and is subject to state constraints, e.g., steering angle or velocity limits.  We denote the set of such states $\sO \subset \sX$. The task of the robot is to avoid these states to get to a target set denoted by $\sT \subseteq \sX$. We are now ready to define our problem.
\begin{problem}[Motion Planning under Unknown Noise]\label{problem:reach-avoid}
Given the robotic system in \eqref{eq:system_dynamics}, the measurable obstacle and target sets $\sO, \sT \subseteq \sX$ introduced above, unsafety probability thresholds $\varepsilon_\sO, \varepsilon_\sT \in (0, 1)$,
and confidence-level threshold $\beta \in (0,1)$, find a control sequence $\bm{u}_{H-1} \in \sU^{H}$ for some horizon $H\in \mathbb{N}$
    such that
    % \begin{enumerate}
    %     \item for all $k = 1, \dots, H$, $\Prob_{x_k}^{\bm u_{k-1}}(\sO) \leq \varepsilon_\sO$,
    %     \item $\Prob^{\bm u_{H-1}}_{x_H}(\sT) \geq 1 - \varepsilon_\sT$.
    % \end{enumerate}
    \begin{align}
        \Prob_{x_k}^{\bm u_{k-1}}(\sO) \leq \varepsilon_\sO
        \quad \forall  k \in [H],
        \quad \text{and}
        \quad
        \Prob^{\bm u_{H-1}}_{x_H}(\sT) \geq 1 - \varepsilon_\sT
    \end{align}
    hold with probability at least $1-\beta$.
\end{problem}

The above chance-constrained motion planning problem is particularly challenging since System~\eqref{eq:system_dynamics} is both nonlinear and stochastic, and the distributions $\Prob_0$ and $\Prob_\omega$ are unknown. To address these challenges, our approach, introduced in Section~\ref{section:learn-to-propagate-method}, combines techniques from distribution learning \cite{fournier2015rate, weed2019sharp, figueiredo2026efficient}, uncertainty propagation \cite{figueiredo2025efficient}, and distributionally robust optimization \cite{gao2023distributionally,figueiredo2025efficient}
with sampling-based planning algorithms (e.g., RRT).
%The overall framework is illustrated in Figure \citehere \LL{Should we add a figure illustrating the overall pipeline, where the RRT nodes correspond to ambiguity sets of distributions and the planner follows a multi-stage propagation scheme?}. 
The key novelty is learning Wasserstein ambiguity sets from data and propagating them through nonlinear dynamics within a search algorithm. This enables us to formally account for the uncertainty induced by finite data while propagating, through the system dynamics, sets of distributions that are statistically consistent with the observed data.
In Section \ref{subsection:propagating}, we introduce the uncertainty propagation framework used to evolve these ambiguity sets over time. Our full planning algorithm is then presented in Section \ref{subsection:motion-planning-algo}, where we also propose a multi-tree strategy that mitigates the accumulation of uncertainty over long horizons while maintaining soundness guarantees.

\section{Learn-To-Propagate Motion Planning from Data}\label{section:learn-to-propagate-method}

Before presenting the full algorithm, we first introduce its core building blocks: the learning and propagation of Wasserstein ambiguity sets, which drive the tree expansion mechanism and are, in principle, applicable to most kinodynamic sampling-based motion planners.

\subsection{Learning Wasserstein Ambiguity Sets}\label{subsection:learning}
Our approach is based on learning Wasserstein ambiguity sets centered at discrete distributions. As shown below, this representation allows us to leverage existing frameworks for learning Wasserstein ambiguity sets while also enabling particularly efficient propagation through the system dynamics. %In this section, we introduce how one can learn a generic distribution $\Prob$ from i.i.d. samples. 

Given a set of $N+N_\text{pre}$ i.i.d. samples $\sD_{N+N_\text{pre}}$ from $\Prob$, our goal is to construct a discrete distribution $\ProbHat$, together with a radius $\varepsilon \geq 0$, such that, with high probability, $\Prob \in \ball_{\varepsilon}(\ProbHat)$. To this end, we follow an approach inspired by \cite{figueiredo2026efficient}, extending it to distributions with unknown and potentially unbounded support. 
% Our method, denoted $\texttt{Learn}(\Prob,\beta)$,
The method is as follows.
We first separate $\sD_{N+N_\text{pre}}$ into a pre-training dataset $\sD_{N_\text{pre}}$ and a training dataset $\sD_{N}$. From $\sD_{N_\text{pre}}$, we compute its bounding box (i.e., an axis-aligned hyperrectangle containing all of its samples) and partition it into $M$ axis-aligned hyperrectangles, $M\in\natNum$ being the desired support size, by cyclic bisections\footnote{For simplicity, we use (uniform) cyclic bisection because it allows us to easily control the size of the cells, which is needed for proving probabilistic completeness. In practice, any partitioning strategy, such as $k$-means, also works, with the caveat that controlling the size of the induced subsets may be harder.}: $\lfloor\log_2M\rfloor$ rounds halve every current cell along the coordinates in cyclic order, and a last round halves $M-2^{\lfloor\log_2M\rfloor}$ of the resulting cells (none if $M$ is a power of two) along the next coordinate. The sets $C_1,\dots,C_M$ are the intersections of these hyperrectangles with $\sX$, and the support of the discrete distribution $\ProbHat$ is $\{ c_1,\dots,c_M \}$, where $c_i\in C_i$ is the center of the $i$-th hyperrectangle if it belongs to $\sX$, and an arbitrary point of $C_i$ otherwise.\footnote{The values of $M$ are given in Appendix~\ref{section:appendix-experiments-details}.
 %Appendix \ref{section:appendix-algorithms}.
 }
The following theorem then provides a method for constructing
(i) a nominal discrete distribution $\ProbHat$ by counting the number of observed samples from $\sD_N$ in each region, and
(ii) a Wasserstein ambiguity set $\ball_{\varepsilon}(\ProbHat)$  such that $\Prob|_{\sX_c} \in \ball_{\varepsilon}(\ProbHat)$, where $\sX_c=\cup_{i=1}^{M} C_i$, with guarantees.\footnote{We stress that, although the ambiguity set is constructed only over the bounded-by-construction region $\sX_c$, Proposition~\ref{prop:extend-to-non-compact} shows that this is sufficient to recover probabilistic guarantees over the full space $\sX$ by controlling the probability mass outside $\sX_c$.}
% 
% The distribution $\ProbHat$ is then constructed in Theorem \ref{thm:learning-ambiguity-ball} below by counting the number of observed samples from $\sD_N$ in each region. Specifically, 
% Theorem \ref{thm:learning-ambiguity-ball} returns a Wasserstein ambiguity set for the conditional distribution $\Prob|_{\sX_c}$, where $\sX_c=\cup_{i=1}^M C_i$. 
% 
% We stress that, although the ambiguity set is constructed only over the bounded-by-construction region $\sX_c$, Proposition~\ref{prop:extend-to-non-compact} later shows that this is sufficient to recover probabilistic guarantees over the full space $\sX$ by controlling the probability mass outside $\sX_c$.
% 
\begin{theorem}[Learning Ambiguity Set]
    \label{thm:learning-ambiguity-ball}
    Let $\sD_N = \{ x_1,\dots,x_N \}$ be i.i.d. samples from $\Prob$ and $\beta \in (0, 1)$ be a confidence level threshold. For $i \in [M]$, let  $Z_i = \sum_{n=1}^{N}  \indicator_{C_i}(x_n)$ be the number of samples from $\sD_N$ in $C_i$ and $N_{\sX_c} =\sum_{i=1}^{M}  Z_i$. Further, define the discrete distribution $\ProbHat = \sum_{i=1}^{M} \frac{Z_i}{N_{\sX_c}} \delta_{c_i}$ associating probability $\frac{Z_i}{N_{\sX_c}}$ to each support point $c_i$. For $i\in [M]$, define lower- and upper-bound probabilities $\evlowerb{i}, \evupperb{i}$, where $\evlowerb{i}=0$ if $Z_i=0$, otherwise $\evlowerb{i}$ is such that
    \[
            \frac{\beta}{2M}=\sum_{j=Z_i}^{N_{\sX_c}} \binom{N_{\sX_c}}{j}(\evlowerb{i})^j(1-\evlowerb{i})^{N_{\sX_c}-j},
    \]
    and $\evupperb{i}=1$ if $Z_i=N_{\sX_c}$, otherwise $\evupperb{i}$ is such that
    \[
            \frac{\beta}{2M}=\sum_{j=0}^{Z_i} \binom{N_{\sX_c}}{j}(\evupperb{i})^j(1-\evupperb{i})^{N_{\sX_c}-j}.
    \]
    Finally, let $\epsilon(\samplesVector)$ be the solution of the following mixed integer linear problem (MILP):
    \begin{align}\label{eq:def-epsilon}
    \epsilon(\samplesVector)^{\rho}&= \sup_{\substack{
    \omega \in \realNum^{M}_{+}, \\
    \gamma \in \realNum^{M \times M}_{+}
    }} \sum_{i=1}^{M} \sum_{j=1}^{M} \max_{x \in C_i} \norm{x-c_j}^{\rho} \evgamma{i,j} \\
    &\qquad \text{s.t. }\;
\left\{
\begin{array}{l}
\nonumber
\gamma \bar\indicator = \omega \\
\gamma^\top \bar\indicator = (Z_1/N_{\sX_c},\dots,Z_{M}/N_{\sX_c}) \\
\evgamma{i,i} \geq \min \bigl\{ \evomega{i}, Z_i/N_{\sX_c} \bigr\},
\quad \forall i \in [M] \\
\evomega{i} \in [\evlowerb{i}, \evupperb{i}],
\quad \forall i \in [M]
\end{array}
\right.
\end{align}
Then, with probability at least $1-\beta$, it holds that $\Prob|_{\sX_c} \in \ball_{\epsilon(\samplesVector)}(\ProbHat)$.
\end{theorem}

Theorem~\ref{thm:learning-ambiguity-ball} generalizes Theorem~3.1 in \cite{figueiredo2026efficient} by allowing the support of $\Prob$ to be unknown and potentially unbounded. Importantly, Proposition~\ref{prop:extend-to-non-compact} below, which is a direct consequence of the law of total probability, guarantees that estimating $\Prob|_{\sX_c}$ allows one to control the probability of arbitrary measurable sets for $\Prob$. All proofs are provided in Appendix~\ref{section:appendix-proofs}.

\begin{proposition}[Validity Check]
    \label{prop:extend-to-non-compact}
    Consider the setting in Theorem \ref{thm:learning-ambiguity-ball}. Let $\tilde\beta \in (0,1)$ be a confidence level threshold and the random variable $N_{\bar{\sX}_c} = N - N_{\sX_c}$ be the number of samples outside $\sX_c$, and define $p_\ell, p_u$ as respectively satisfying
$\frac{\tilde\beta}{2}=\sum_{j=N_{\bar{\sX}_c}}^{N} \binom{N}{j} p_\ell^j (1-p_\ell)^{N-j}$ and $\frac{\tilde\beta}{2}=\sum_{j=0}^{N_{\bar{\sX}_c}} \binom{N}{j} p_u^j (1-p_u)^{N-j}$.\footnote{As in Theorem~\ref{thm:learning-ambiguity-ball}, we set $p_\ell=0$ if $N_{\bar{\sX}_c}=0$ and $p_u=1$ if $N_{\bar{\sX}_c}=N$.}
    Then, for any set $\sS \in \sB(\sX)$, it holds with probability at least $1-\beta-\tilde\beta$ that
    \begin{equation}
    \label{eq:Probability}
        \Prob(\sS) \leq (1-p_\ell) \cdot \sup_{ \ProbQ \in \ball_{\epsilon(\sD_N)}(\ProbHat) } \ProbQ(\sS) + p_u.
    \end{equation}
\end{proposition}
In Proposition~\ref{prop:extend-to-non-compact}, the interval $(p_\ell,p_u)$ is constructed to bound $\Prob(\sX\setminus\sX_c)$ with high confidence. We refer to it as the ambiguity set's \emph{escaped mass}. 
Note that the bound in \eqref{eq:Probability} still requires to compute $\sup_{ \ProbQ \in \ball_{\epsilon(\sD_N)}(\ProbHat) } \ProbQ(\sS)$. While this is not trivial in general, in our case, where $\ProbHat$ is discrete with support size $M$, this can be done particularly efficiently in time $O(M\log M)$ by sorting the distance of each support point of $\ProbHat$ to the set $\sS$, and then greedily moving mass to this set until the Wasserstein distance constraint $\epsilon(\sD_N)$ is met \cite{gao2023distributionally}.

% These results give rise to two procedures: 
% (i) $\texttt{LearnAmbiguitySet}(\Prob, \beta)$ that constructs ambiguity set $\ball_{\epsilon}(\ProbHat)$ that contains $\Prob$ with probability at least $1-\beta$, and
% (ii) $\texttt{SafetyCheck}(\ball_\epsilon(\ProbHat),(p_\ell, p_u),\varepsilon)$ that checks whether $\Prob(\sS) \leq \varepsilon$ with probability at least $1-\beta$. 
% See the supplementary material for the pseudocodes.

These results give rise to two procedures:
(i) $\texttt{LearnAmbiguitySet}(\sD, \beta, M)$, taking a dataset $\sD=\sD_{N+N_\text{pre}}$ of i.i.d.\ samples from $\Prob$ and a support size $M$, and constructing, with probability at least $1-\beta$, an ambiguity set $\ball_{\epsilon}(\ProbHat)$ (with $|\mathrm{supp}(\ProbHat)|= M$) containing $\Prob|_{\sX_c}$, and escaped-mass bounds $p_\ell, p_u$ such that $p_\ell \leq \Prob(\sX \setminus \sX_c) \leq p_u$ for a bounded constructed set $\sX_c$; and
(ii) \texttt{ValidityCheck}, which takes a set $\sS$ and a threshold $\varepsilon$, and certifies $\Prob(\sS) \leq \varepsilon$ with probability at least $1-\beta$
(see Appendix~\ref{section:appendix-algorithms}).

% From these results, 
% We provide a summary of the complete learning procedure \texttt{LearnAmbiguitySet} and the procedure for solving of the optimization problem above (\texttt{SafetyCheck}) in the supplementary material.
%in Algorithm \ref{alg:learn-ambiguity} (\texttt{LearnAmbiguitySet}) in Appendix \ref{section:appendix-algorithms}, as well as Algorithm \ref{alg:compute-bounds-in-probability} (\texttt{SafetyCheck}) on how to solve the optimization problem above.

\subsection{Propagating Wasserstein Ambiguity Sets}\label{subsection:propagating}

% To complete the algorithm, we need to propagate the ambiguity sets obtained via Theorem~\ref{thm:learning-ambiguity-ball} (\texttt{LearnAmbiguitySet} procedure) through the system dynamics multiple times. This can be efficiently performed as follows. Consider a Wasserstein ambiguity set $\ball_{\epsilon}(\ProbHat)$ and an arbitrary Lipschitz continuous function $h:\realNum^d \to \realNum^d$  with Lipschitz constant $\sL_h >0$. Applying \cite[Thm~5.1]{figueiredo2025efficient} yields:

To complete the algorithm, we need to propagate the ambiguity sets obtained via Theorem~\ref{thm:learning-ambiguity-ball} (the \texttt{LearnAmbiguitySet} procedure) through the system dynamics over multiple time steps. This can be performed efficiently as follows. Consider a Wasserstein ambiguity set $\ball_{\epsilon}(\ProbHat)$ and an arbitrary Lipschitz continuous function $h:\realNum^d \to \realNum^d$ with Lipschitz constant $\sL_h > 0$. Specializing \cite[Thm~5.1]{figueiredo2025efficient} to this setting yields:
\begin{equation}\label{eq:propagation-statement}
    h \#\ball_{\epsilon}(\ProbHat) \subseteq \ball_{\sL_{h}\epsilon} \big( h \# \ProbHat \big).
\end{equation}
% Critically, when center distribution $\ProbHat$ is discrete, the propagated center distribution $h\# \ProbHat$ can be computed efficiently. Indeed, by denoting $\ProbHat = \sum_{i=1}^M \evpi{i} \delta_{c_i}$, where $\pi \in [0,1]^{M}$ is a probability vector, we have that $h \# \ProbHat = \sum_{i=1}^M \evpi{i}\delta_{h(c_i)}$, i.e., each support point $c_i$ is simply mapped to $h(c_i)$. 
Critically, when the center distribution $\ProbHat$ is discrete, the propagated center $h \# \ProbHat$ can be computed efficiently. Indeed, writing $\ProbHat = \sum_{i=1}^{M} \evpi{i} \delta_{c_i}$ for a probability vector $\pi \in [0,1]^{M}$, we have $h \# \ProbHat = \sum_{i=1}^{M} \evpi{i}\delta_{h(c_i)}$; that is, each support point $c_i$ is mapped to $h(c_i)$.
% We note that, although that theorem provides tighter bounds than the Lipschitz-based estimate, we adopt the Lipschitz bound here for computational tractability. While this choice may conservatively enlarge the propagated ambiguity sets, the conservatism is mitigated in our full algorithm by periodically relearning the uncertainty, as discussed in Section \ref{section:learn-to-propagate-method}.
% Algorithm \texttt{Propagate}, presented in the supplementary material,
%~\ref{alg:generate-node} (\texttt{GenerateNode}), presented in Appendix \ref{section:appendix-algorithms}, 
% describes the core mechanism driving tree expansion in our motion planning algorithm, and it is built on the specialization of \eqref{eq:propagation-statement} to our setting by considering $h = f(\cdot, u, \cdot)$ for a randomly selected control $u \in \sU$, and the learned ambiguity $\ball_{\epsilon}(\ProbHat_x \otimes \ProbHat_\vomega)$.   
 We note that the Lipschitz-based propagation method in \eqref{eq:propagation-statement} is significantly more computationally tractable than the general result in \cite[Thm~5.1]{figueiredo2025efficient}. This efficiency is important for fast exploration of the search space in motion planning. However, it may yield conservatively large propagated ambiguity sets. Our planning algorithm uses a multi-tree (forest) strategy to mitigate this conservatism as described in Section~\ref{subsection:motion-planning-algo}. 
 
These results lead to the procedure \texttt{Propagate} (see pseudocode in Appendix~\ref{section:appendix-algorithms}), the core mechanism driving tree expansion in our motion planning algorithm. Given an ambiguity set, state and noise escaped masses $(p_{\ell},p_{u}), (p_{\vomega,\ell},p_{\vomega,u})$, and a target Delta-dirac $\delta_x$ centered at a given state $x$, Propagate picks a control $u \in \sU$ to steer that ambiguity set's center distribution towards $\delta_x$, and returns the propagated Wasserstein ambiguity set obtained through $h = f(\cdot, u, \cdot)$ via \eqref{eq:propagation-statement}, as well as the escaped mass via Proposition~\ref{prop:specialize-prop-to-problem}.

\iffalse
\begin{remark}[Additive noise]
    In particular, when $f_u(x, \vomega) = g_u(x) + h_u(\vomega)$, \eqref{eq:propagation-statement} (and correspondingly line~\ref{step:propagate-radius} in Algorithm~\ref{alg:generate-node}) can be simplified to
    \begin{equation}\label{eq:propagation-statement-separable-case}
        f \# \big( \ball_{\epsilon_x}(\ProbHat_x ) \otimes \ball_{\epsilon_\vomega}(\ProbHat_\vomega ) \big) \subseteq \ball_{\sL_{g_u}\epsilon_x+\sL_{h_u}\epsilon_\vomega} \big( g_u \# \ProbHat_x * h_u\#\ProbHat_\vomega \big),
    \end{equation}
    where $\sL_{g}, \sL_h$ are the Lipschitz constants of $g, h$ and $*$ represents the convolution operator. 
\end{remark}
\fi

\subsection{Learn-To-Propagate Distributionally-Robust Forest (L2P-DR-Forest) Algorithm}\label{subsection:motion-planning-algo}
We are now ready to present our motion planning algorithm, called \textit{Learn-To-Propagate Distributionally-Robust Forest} (L2P-DR-Forest), whose pseudocode is given in Algorithm~\ref{alg:learn-to-propagate-rrt}.
% 
% We note that, 
% having defined procedures \texttt{LearnAmbiguitySet}, \texttt{ValidityCheck}, and \texttt{Propagate}, we can adapt any off-the-shelf tree-based kinodynamic sampling-based motion planner in our framework.  
% For clarity and proof of correctness,
% specifically we describe the method using RRT as the base planner.
% 
% 
% RRT motion planning algorithm, described in Algorithm~\ref{alg:learn-to-propagate-rrt}. 
% Note that, with the defined procedures \texttt{LearnAmbiguitySet}, \texttt{ValidityCheck}, and \texttt{Propagate}, the proposed framework applies to any tree-based kinodynamic sampling-based motion planners, but we describe the method using RRT as the base planner for concreteness.
% 
% The algorithm 
It operates in stages and grows one motion tree per stage, which yields a forest. Within each stage, the tree is expanded according to standard tree-based kinodynamic sampling-based motion planning, e.g., RRT. The forest differs from existing multi-tree planners~\cite{Tin2022multitrees, Tin2019randomdisjointtrees, Plaku2005roadmapoftrees},
which grow trees from distinct states and connect them. Instead, we root each new tree at a node of an earlier tree and re-learn its
ambiguity set from new data.

\begin{figure*}[t]
    \centering
    \includegraphics[width=\textwidth]{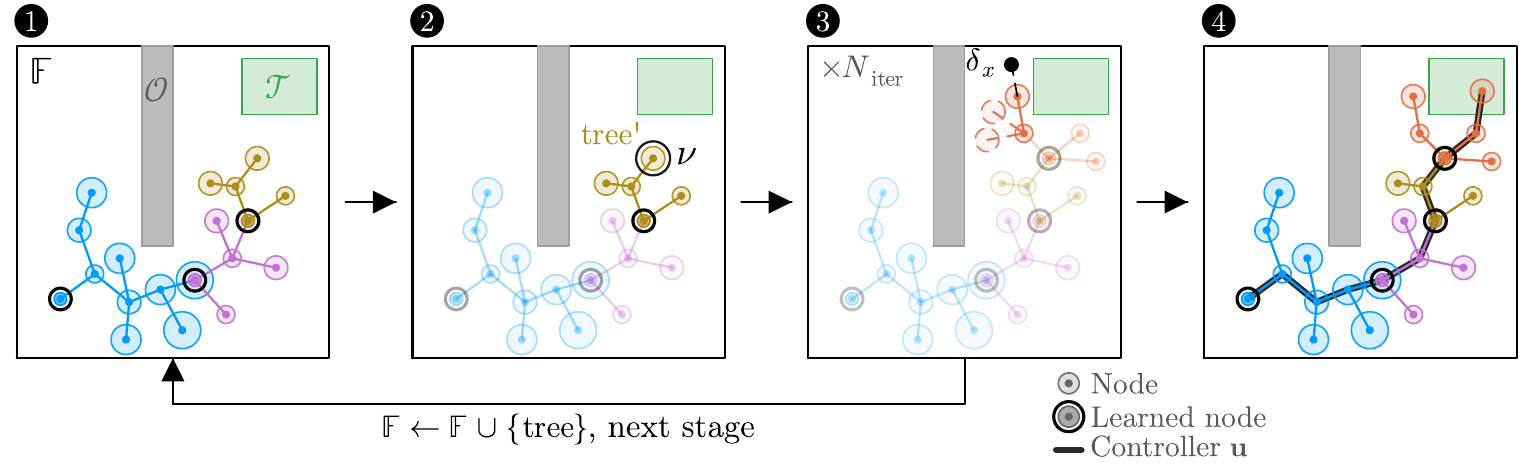}
    \caption{Schematic of Algorithm~\ref{alg:learn-to-propagate-rrt} on a scenario with an obstacle $\sO$ and a target $\sT$. The roots of the trees, learned with \texttt{LearnAmbiguitySet}, are highlighted. \emph{(1)}~The forest $\mathbb{F}$ stores the trees of the previous stages. \emph{(2)}~\texttt{SelectRoot} selects one of the available trees in $\mathbb{F}$, and a node $\nu$ (circled) from this tree, at which the root of the new tree is learned (Lines~\ref{step:select-root}--\ref{step:learn-noise}). \emph{(3)}~Each iteration of the tree search uniformly samples a state $x \in \sX$ (black point), selects the node closest to it, and steers it towards $x$ with the best of $K$ candidate controls (solid child; the others are dashed) (Lines~\ref{step:expand-select}--\ref{step:add-node}). If no node passes \texttt{TargetCheck}, the tree is added to $\mathbb{F}$ and the next stage begins (Line~\ref{step:add-forest}). \emph{(4)}~Otherwise, \texttt{backtrack} returns the controller $\bm{u}$ leading to the node that passes (black path; Lines~\ref{step:target-check}--\ref{step:return}).}
    \label{fig:algorithm-schema}
\end{figure*}

The trees are stored in a forest $\mathbb{F}$, initially empty, and each stage has four steps: (i) it selects a tree in $\mathbb{F}$ and a node $\nu$ from it (Line~\ref{step:select-root}); (ii) it learns, from fresh data, the ambiguity set of the state distribution induced by the controller leading to $\nu$ (Lines~\ref{step:roll-forward}--\ref{step:learn-noise}); (iii) it grows a new tree rooted at the newly learned node (Lines~\ref{step:init-tree}--\ref{step:return}); and (iv) it adds the tree to the forest (Line~\ref{step:add-forest}). Figure~\ref{fig:algorithm-schema} illustrates our proposed algorithm.  The nodes of the trees represent ambiguity sets, learned, propagated, and checked using the methods described in the previous subsections.
Specifically, the root node of each tree is a Wasserstein ambiguity set that is learned directly from data using \texttt{LearnAmbiguitySet}(), and all of its offspring nodes are propagated using \texttt{Propagate}(). The noise distribution $\Prob_\vomega$ is learned in the same way, once per stage. In other words, \texttt{LearnAmbiguitySet}() is used only twice per stage: once for the root of its tree and once for the noise.
Each node stores: (i) a Wasserstein ambiguity set, together with its escaped-mass bounds $(p_\ell,p_u)$,
(ii) the control action used to reach it (for a root, the controller $\bm u$ leading to it), and
(iii) a pointer to its parent node.

\emph{Restart node.} At the beginning of each stage, $\texttt{SelectRoot}(\mathbb{F};\lambda_1,\lambda_2)$ selects the restart node $\nu$ (Line~\ref{step:select-root}) in two steps. First, it selects a tree $\texttt{tree}'$ of the forest: the most recent one with probability $1-\lambda_1$, and one drawn uniformly from $\mathbb{F}$ otherwise. Then, it selects in $\texttt{tree}'$ the restart node $\nu$: with probability $1-\lambda_2$ the promising node given by the heuristic of Appendix~\ref{subsection:select-promising-trajectory}, and a node drawn uniformly from $\texttt{tree}'$ otherwise. Such exploration (by setting both $\lambda_1, \lambda_2>0$) prevents the algorithm from committing irreversibly to a single refined prefix and will prove to be fundamental to show the probabilistic completeness of our algorithm in Theorem~\ref{thm:probabilistic-completeness}. The stage then restarts at the controller $\bm u=\texttt{backtrack}(\nu)$ leading to $\nu$ (Line~\ref{step:backtrack}).

\emph{Learning.} The datasets $\sD_{x_0}, \sD_\vomega$ supply not-yet-used, mutually disjoint subsets for the stage: $\sD_{x_0}^{(s)}, \sD_\vomega^{(s)}$, which are rolled forward under $\bm u$ into a dataset $\sD_{N(s)}^{(s)}$ of state samples (Line~\ref{step:roll-forward}), and a batch $\sD_{\vomega,N_\vomega(s)}^{(s)}$ of $N_\vomega(s)$ noise samples. Formally, for a controller $\bm u=(u_0,\dots,u_{k-1})$, an initial state $x_0$ and a noise sequence $\bm\vomega=(\vomega_0,\dots,\vomega_{k-1})$, the state reached at time $k$ is
\begin{equation}\label{eq:roll-forward}
    F_{\bm u}(x_0,\bm\vomega)\coloneqq x_k, \qquad x_{j+1}=f(x_j,u_j,\vomega_j), \quad j=0,\dots,k-1,
\end{equation}
and Line~\ref{step:roll-forward} applies \eqref{eq:roll-forward} with $k=|\bm u|$ to the $i$-th initial state of $\sD_{x_0}^{(s)}$ and the $i$-th block of $|\bm u|$ consecutive samples of $\sD_\vomega^{(s)}$, for $i=1,\dots,N(s)$. The $N(s)$ trajectories use $|\bm u|$ fresh noise samples each, one per step, i.e., $N(s)|\bm u|$ in total, so that $\sD_{N(s)}^{(s)}$ is i.i.d.\ from the state distribution induced by $\bm u$. Since the length of the stage controller $|\bm u|$ is not known a priori, the stage partition $\sD_{\vomega,N_\vomega(s)}^{(s)}$ of the noise dataset $\sD_{\vomega,N_\vomega(s)}$ must be performed during runtime, while $\sD_{x_0}^{(s)}$ can be defined prior to execution. \texttt{LearnAmbiguitySet}() invokes Theorem~\ref{thm:learning-ambiguity-ball} on these two datasets to construct a high-confidence Wasserstein ambiguity set and escaped-mass bound for the root (Line~\ref{step:learn-root}) and for the noise (Line~\ref{step:learn-noise}). The confidence schedule $(\beta_s)_{s=1}^{N_\text{stage}}$ allocates a share $\beta_s$ of the overall confidence budget $\beta$ to stage $s$: $\zeta\beta_s$ for the root's state distribution learning and $(1-\zeta)\beta_s$ for the noise. Any schedule with $\sum_{s=1}^{N_\text{stage}}\beta_s\leq\beta$ is valid\footnote{For instance, the uniform schedule $\beta_s=\beta/N_\text{stage}$ when $N_\text{stage}$ is finite, the polynomial schedule $\beta_s=6\beta/(\pi^2s^2)$, or the exponential schedule $\beta_s=\beta(1-\gamma)\gamma^{s-1}$ for any fixed $\gamma\in(0,1)$, for any (possibly infinite) $N_\text{stage}$.}. The schedules $(N(s),M(s))$ and $(N_\vomega(s),M_\vomega(s))$ set the sample size and the support size of the root and noise calls, respectively.

\emph{Tree search.} A new tree is initialized at the learned root node (Line~\ref{step:init-tree}), and
the tree search routine (Lines~\ref{step:expand-loop}--\ref{step:return}) expands it in a sampling-based fashion, e.g., RRT, for $N_{\text{iter}}$ iterations, after which it is added to the forest $\mathbb{F}$ (Line~\ref{step:add-forest}). In Lines~\ref{step:expand-select}--\ref{step:propagated-node}, \texttt{SelectNode}() selects a node\footnote{In case of RRT, a random distribution can be sampled by sampling $x_\text{rand}$ and defining distribution $\delta_{x_\text{rand}}$. The nearest neighbor is computed according
to the distance $\norm{x_\text{rand}-\bar x_i}$ between $x_\text{rand}$ and the mean $\bar x_i$ of $\ProbHat_i$, the center of the ambiguity set of node $i$ (Appendix~\ref{section:appendix-algorithms}). Ties are broken uniformly at random.},
and \texttt{Propagate}() propagates the ambiguity set and escaped-mass bound of the selected node using \eqref{eq:propagation-statement}  and Proposition~\ref{prop:specialize-prop-to-problem}, respectively, under a chosen control input $u^\star \in \sU$. This new node is added to the tree only if it satisfies the distributionally robust safety condition of Problem~\ref{problem:reach-avoid} (Lines~\ref{step:validity-check}--\ref{step:add-node}), and the algorithm terminates successfully once a node satisfies the distributionally robust target condition, returning the controller leading to it (Lines~\ref{step:target-check}--\ref{step:return}).

\RestyleAlgo{ruled}
\SetKwComment{Comment}{/* }{ */}
\begin{algorithm*}[t]
\DontPrintSemicolon
\caption{Learn-To-Propagate Distributionally-Robust Forest (L2P-DR-Forest)}
\label{alg:learn-to-propagate-rrt}
\KwIn{Datasets $\sD_{x_0}, \sD_\vomega$, unsafety thresholds $\varepsilon_\sT, \varepsilon_\sO \in (0, 1)$, number of stages $N_{\texttt{stage}}>0$, number of iterations $N_{\texttt{iter}} > 0$, confidence level $\beta \in (0,1)$, confidence weight $\zeta\in(0,1)$, confidence schedule $(\beta_s)_{s=1}^{N_\text{stage}}$ with $\sum_{s=1}^{N_\text{stage}}\beta_s\leq\beta$, size schedules $(N(s), M(s))_{s=1}^{N_\text{stage}}$ and $(N_\vomega(s), M_\vomega(s))_{s=1}^{N_\text{stage}}$, exploration probabilities $\lambda_1,\lambda_2\in[0,1)$.}
\KwOut{Solution $\bm{u}_{H-1}$ to Problem \ref{problem:reach-avoid}.}

Initialize $\mathbb{F} \gets \emptyset$

\For{$s \gets 1$ \KwTo $N_{\textup{\texttt{stage}}}$}{

    $\nu \gets \texttt{SelectRoot}(\mathbb{F}; \lambda_1, \lambda_2)$ \label{step:select-root}

    $\bm u \gets \texttt{backtrack}(\nu)$ \label{step:backtrack}

    $\sD_{N(s)}^{(s)} \gets \big\{ F_{\bm u}(x_0,\bm\vomega) : (x_0,\bm\vomega) \in \sD_{x_0}^{(s)} \times \sD_\vomega^{(s)} \text{ paired by index} \big\}$ \label{step:roll-forward}

    $\ball_{\epsilon_\text{root}}(\ProbHat_\text{root}), (p_\ell, p_u) \gets \texttt{LearnAmbiguitySet}\big(\sD_{N(s)}^{(s)}, \zeta\beta_s, M(s) \big)$ \label{step:learn-root}

    $\ball_{\epsilon_\vomega}(\ProbHat_{\vomega}), (p_{\vomega, \ell}, p_{\vomega, u}) \gets \texttt{LearnAmbiguitySet}\big(\sD_{\vomega,N_\vomega(s)}^{(s)}, (1-\zeta)\beta_s, M_\vomega(s) \big)$ \label{step:learn-noise}

    $\texttt{tree} \gets \big\{ \texttt{node}_\text{root} \gets \big( \ball_{\epsilon_\text{root}}(\ProbHat_\text{root}), (p_\ell,p_u), \bm u, \varnothing\big) \big\}$ \label{step:init-tree}

    \For{$i \gets 1$ \KwTo $N_{\textup{\texttt{iter}}}$ \label{step:expand-loop}}{

        $\texttt{node}_\text{selected}, \delta_x \gets \texttt{SelectNode}(\texttt{tree})$ \label{step:expand-select}

        $\big( \ball_\epsilon(\ProbHat), (p_\ell', p_u'), u^\star \big) \gets \texttt{Propagate}\big(\texttt{node}_\text{selected}, \ball_{\epsilon_\vomega}(\ProbHat_{\vomega}), (p_{\vomega,\ell},p_{\vomega,u}), \delta_x\big)$ \label{step:propagate-call}

        $\texttt{node}_\text{propagated} \gets \big( \ball_\epsilon(\ProbHat), (p_\ell', p_u'), u^\star, \texttt{node}_\text{selected} \big)$ \label{step:propagated-node}

        \If{$\emph{\texttt{ValidityCheck}}\big( \texttt{node}_\text{propagated}, \mathcal{O}, \varepsilon_\sO \big)$ \label{step:validity-check}}{
            $\texttt{tree} \gets \texttt{tree} \cup \{ \texttt{node}_\text{propagated} \}$ \label{step:add-node}\;
            \If{$\emph{\texttt{TargetCheck}}\big( \texttt{node}_\text{propagated}, \mathcal{T}, \varepsilon_\sT \big)$ \label{step:target-check}}{
                \Return $\bm{u} \gets \texttt{backtrack}\big( \texttt{node}_\text{propagated} \big)$ \label{step:return}\;
            }
        }
    }

    $\mathbb{F} \gets \mathbb{F} \cup \{\texttt{tree}\}$ \label{step:add-forest}
}
\end{algorithm*}

The computational cost of each tree expansion is dominated by the (robust) \texttt{ValidityCheck}() and \texttt{TargetCheck}() procedures. As explained in Section~\ref{subsection:propagating}, propagation itself is efficient because the center distributions are discrete. However, repeated propagation using the Lipschitz-based method may inflate the ambiguity radius, unless dynamics are contractive ($\sL_h < 1$). Consequently, nodes deep in a single tree may carry overly conservative ambiguity sets, causing them to fail the validity check in challenging regions, e.g., narrow passages. 
A key observation is that the ambiguity sets are tightest near the regions where they are directly learned from data.
% , and it is the Lipschitz-based propagation that 
% As an ambiguity set is propagated farther through the dynamics, its radius is enlarged relatively conservatively using the Lipschitz constants as in \eqref{eq:propagation-statement}. 
% Thus, nodes close to a learned or refined root carry relatively tight ambiguity sets, whereas nodes farther away from that root become increasingly conservative.
% 
Our forest approach exploits this locality. At each stage, the root of the new tree is learned afresh from the state distribution induced by the controller $\bm u=\texttt{backtrack}(\nu)$ leading to the restart node $\nu$ (Lines~\ref{step:roll-forward}--\ref{step:learn-root}), rather than reusing the ambiguity set of $\nu$, which may have been inflated by repeated propagation. The nodes of the new tree are propagated from this tight root, so their ambiguity sets stay small enough to pass the validity check farther along the trajectory, e.g., through narrow passages.

\subsubsection{Correctness and Probabilistic Completeness}
The following theorem proves the soundness of the returned solution by Algorithm~\ref{alg:learn-to-propagate-rrt}. 
\begin{theorem}[Correctness of Alg.~\ref{alg:learn-to-propagate-rrt}]\label{thm:correctness-algorithm-learn-to-propagate}
    If Algorithm \ref{alg:learn-to-propagate-rrt} returns a controller $\bm{u}_{H-1}$ (for the corresponding horizon $H$), then $\bm{u}_{H-1}$ solves Problem \ref{problem:reach-avoid} with confidence at least $1-\beta$ over the datasets $\sD_{x_0}, \sD_\vomega$.
\end{theorem}

The proof relies on the fact that, in each stage,  the calls of \texttt{LearnAmbiguitySet} return ambiguity sets and escaped masses containing the true state and noise distributions with probability $1-\beta_s$. Because the propagation and validity checks rely on guaranteed over- and under-approximations, their guarantees hold under the event that the learner is correct. The conclusion thus follows trivially from the application of the union bound and $\sum_{s} \beta_s \leq \beta$.

We now move to showing that Algorithm~\ref{alg:learn-to-propagate-rrt} is probabilistic complete under some mild distributional assumptions. Completeness is usually formalized by requiring a noise-free trajectory to clear the unsafe and non-target regions by a margin $\Delta$~\cite{lavalle2001randomized}. In our stochastic setting, we state it directly on the propagated distributions $\Prob^{\bm u_{k-1}}_{x_k}$ induced by a control sequence. Assumption~\ref{assumption:robust-feasibility} below requires the existence of a nominal controller $\tilde{\bm{u}}$ such that the induced state distributions stay strictly safe under any perturbation of Wasserstein size $\Delta$, thus allowing for a margin in probability space.

\begin{assumption}[$\Delta$-Robust Feasibility]\label{assumption:robust-feasibility}
There exist $\Delta>0$, a horizon $\bar H\in\natNum$ and a control sequence $\tilde{\bm u}_{\bar H-1}=(\tilde u_0,\dots,\tilde u_{\bar H-1})\in\sU^{\bar H}$ such that
\begin{equation}
\begin{aligned}
&\sup_{\ProbQ \in \ball_\Delta\big(\Prob^{\tilde{\bm u}_{k-1}}_{x_k}\big)} \ProbQ(\sO) < \varepsilon_\sO \quad \forall k \in [\bar H], \\
&\sup_{\ProbQ \in \ball_\Delta\big(\Prob^{\tilde{\bm u}_{\bar H-1}}_{x_{\bar H}}\big)} \ProbQ(\sX \setminus \sT) < \varepsilon_\sT.
\end{aligned}
\end{equation}
\end{assumption}
We prove the probabilistic completeness of Algorithm~\ref{alg:learn-to-propagate-rrt} assuming the following limit conditions: the algorithm is run with $N_\text{stage}=\infty$ on infinite i.i.d.\ sequences $\sD_{x_0},\sD_\vomega$, with schedules $\beta_s$, $N(s)$, $M(s)$, $N_\vomega(s)$ and $M_\vomega(s)$ satisfying the conditions of Appendix~\ref{subsection:proof-completeness}.

\begin{theorem}[Probabilistic Completeness]\label{thm:probabilistic-completeness}
Suppose Assumption~\ref{assumption:robust-feasibility} holds, and the execution conditions from the previous paragraph. Further, assume $\Prob_\vomega$ is sub-Gaussian, $\sU$ has positive Lebesgue measure in every ball centered at its points, and Algorithm~\ref{alg:learn-to-propagate-rrt} is run as above with $\lambda_1,\lambda_2\in(0,1)$. Then, with probability $1$, Algorithm~\ref{alg:learn-to-propagate-rrt} returns a controller $\bm{u}_{H-1}$ within finitely many stages that solves Problem~\ref{problem:reach-avoid} with confidence at least $1-\beta$.
\end{theorem}

The proof, provided in Appendix~\ref{section:appendix-proofs},
combines convergence of the learned uncertainty bounds with
persistent exploration.
As the sample sizes increase through successive learning and
re-learning stages, the learned radii and escaped masses converge
to zero.
The sub-Gaussian assumption on $\Prob_\vomega$ ensures that the bounding box of the pre-training samples, which scales the learned radius, grows at most logarithmically with the sample size, so that the radius can vanish.
The robustness margin $\Delta$ in
Assumption~\ref{assumption:robust-feasibility} therefore ensures
that nodes associated with controllers sufficiently close to the
robustly feasible controller $\tilde{\bm u}$ eventually pass
the validity checks.
Moreover, the restart mechanism, with $\lambda_1,\lambda_2>0$,
ensures that the search repeatedly revisits earlier nodes and
explores alternative control extensions.

\section{Empirical Evaluation}\label{section:experimental-results}
We empirically evaluate \emph{L2P-DR-Forest} on five robotic benchmarks: a 2D $5$-mode \emph{Switched System} \cite{gracia2025efficient}, two 3D vehicle models (\emph{AV Maze} \cite{abate2025arch,Calbert2024}, and \emph{Unicycle Maze}), a 4D \emph{Tractor Trailer} \cite{bos2023mpc} with residual neural-network dynamics, and a 6D \emph{Double Integrator}. For all the experiments, we considered $\rho=2$ with the Euclidean norm, a confidence parameter of $\beta=10^{-4}$ , exploration probabilities $\lambda_1 = \lambda_2 = 10^{-2}$, a confidence weight $\zeta=0.75$, the uniform schedule $\beta_s=\beta/N_\text{stage}$, sample sizes $N(s)=N_\vomega(s)$ growing linearly with $s$, and constant support sizes $M(s)=M_\vomega(s)=4$ (or $2$ for \emph{Tractor Trailer}). Dynamics, environmental details, and the remaining hyperparameters are provided in Appendix~\ref{section:appendix-experiments-details}. All experiments were executed on an Apple M2 Pro CPU with 16GB of RAM.\footnote{Code will be made available upon acceptance.}
%Appendix~\ref{section:appendix-experiments-details}

% Generated by experiments/paper/generate_table.jl (do not edit by hand).
\begin{table*}[t]
\centering
\small
\setlength{\tabcolsep}{6pt}
\resizebox{\ifdim\width>\linewidth\linewidth\else\width\fi}{!}{%
\begin{tabular}{clccccc}
\toprule
 & & \emph{Switched System} & \emph{AV Maze} & \emph{Unicycle Maze} & \emph{Tractor Trailer} & \emph{Double Integrator} \\
 & Dimension & 2D & 3D & 3D & 4D & 6D \\
\midrule
\midrule
\multirow{10}{*}{\rotatebox{90}{\scriptsize\shortstack{L2P-DR-Forest\\(20 trials)}\quad \;\;\;}}
    & Max. number of it. per stage & $1{,}500$ & $3{,}000$ & $5{,}000$ & $1{,}500$ & $3{,}000$ \\
  & Max. number of stages & $10$ & $10$ & $20$ & $20$ & $20$ \\
  & Succ. rate (\%) & $100.0$ & $100.0$ & $100.0$ & $100.0$ & $100.0$ \\
  & Total planning time (\si{\second}) & $0.08 \pm 0.12$ & $1.56 \pm 0.96$ & $2.15 \pm 3.17$ & $55.51 \pm 47.22$ & $1.28 \pm 0.37$ \\
  & MILP time (\si{\second}) & $0.0343 \pm 0.0096$ & $0.0427 \pm 0.0107$ & $0.0919 \pm 0.0554$ & $0.1129 \pm 0.0356$ & $0.0785 \pm 0.0180$ \\
  & Peak mem. cons. (\si{\giga\byte}) & $1.14 \pm 0.00$ & $1.21 \pm 0.03$ & $1.29 \pm 0.07$ & $1.37 \pm 0.06$ & $1.24 \pm 0.05$ \\
  & Effect. re-learning used & $2.40 \pm 0.50$ & $4.85 \pm 1.18$ & $6.95 \pm 3.78$ & $10.10 \pm 3.09$ & $7.15 \pm 1.04$ \\
  & Controller length & $8.70 \pm 1.59$ & $68.35 \pm 23.49$ & $37.25 \pm 19.31$ & $92.50 \pm 25.25$ & $49.40 \pm 9.90$ \\
  & Max. empir. unsafe prob. (\%) & $0.0$ & $0.0$ & $0.0$ & $0.0$ & $0.0$ \\
  & Empir. target prob. (\%) & $100.0$ & $100.0$ & $100.0$ & $100.0$ & $100.0$ \\
\midrule
\multirow{6}{*}{\rotatebox{90}{\scriptsize\shortstack{No re-learning\\(20 trials)}\;}}
   & Max. number of it. per stage & $15{,}000$ & $30{,}000$ & $100{,}000$ & $30{,}000$ & $60{,}000$ \\
  & Max. number of re-learning & $1$ & $1$ & $1$ & $1$ & $1$ \\
  & Succ. rate (\%) & $0.0$ & $0.0$ & $0.0$ & $0.0$ & $0.0$ \\
  & Total planning time (\si{\second}) & -- & -- & -- & -- & -- \\
  & Peak mem. cons. (\si{\giga\byte}) & -- & -- & -- & -- & -- \\
  & Controller length & -- & -- & -- & -- & -- \\
\midrule
\multirow{8}{*}{\rotatebox{90}{\scriptsize\shortstack{$\epsilon$-RandUP-RRT\\(20 trials)} \hspace{-0.2cm} \;}}
    \vspace{-0.2cm} \\
  & Succ. rate (\%) & $0.0$ & $85.0$ & $80.0$ & $0.0$ & $100.0$ \\
  & Total planning time (\si{\second}) & -- & $31.30 \pm 25.60$ & $9.58 \pm 19.17$ & -- & $2.02 \pm 1.14$ \\
  & Number of timeouts (2 min) & $0$ & $3$ & $4$ & $20$ & $0$ \\
  & Controller length & -- & $60.29 \pm 6.79$ & $25.94 \pm 3.92$ & -- & $63.15 \pm 2.01$ \\
  & Max. empir. unsafe prob. (\%) & -- & $0.0$ & $0.0$ & -- & $0.0$ \\
  & Empir. target prob. (\%) & -- & $100.0$ & $100.0$ & -- & $100.0$ \\
  \vspace{-0.2cm} \\
  \bottomrule
\end{tabular}}
\caption{Summary of average selected statistics (with one standard deviation).}
\label{tab:experimental_results_comparison_solved}
\end{table*}

\begin{figure*}[t]
    \centering
    \begin{subfigure}[t]{0.19\textwidth}
        \centering
        \includegraphics[width=\linewidth]{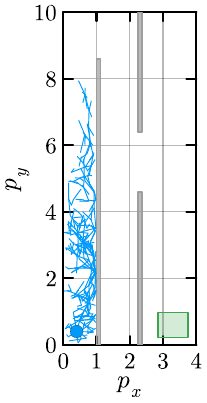}
        \caption{First stage}
        \label{fig:maze_image1}
    \end{subfigure}
    \hfill
    \begin{subfigure}[t]{0.19\textwidth}
        \centering
        \includegraphics[width=\linewidth]{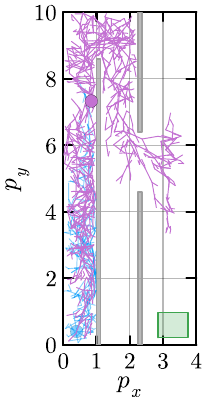}
        \caption{Second stage}
        \label{fig:maze_image2}
    \end{subfigure}
    \hfill
    \begin{subfigure}[t]{0.19\textwidth}
        \centering
        \includegraphics[width=\linewidth]{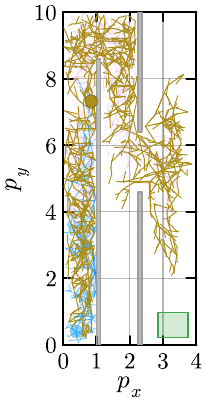}
        \caption{Third stage (restarts in tree 1)}
        \label{fig:maze_image3}
    \end{subfigure}
    \hfill
    \begin{subfigure}[t]{0.19\textwidth}
        \centering
        \includegraphics[width=\linewidth]{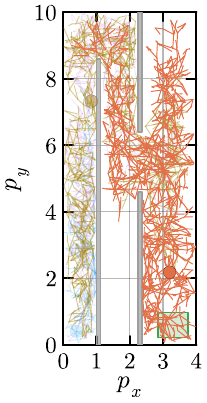}
        \caption{Fourth (last) stage}
        \label{fig:maze_image4}
    \end{subfigure}
    \hfill
    \begin{subfigure}[t]{0.19\textwidth}
        \centering
        \includegraphics[width=\linewidth]{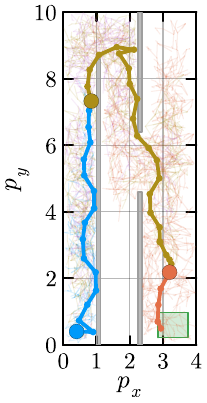}
        \caption{Safe trajectory}
        \label{fig:maze_image5}
    \end{subfigure}
    \caption{\emph{AV Maze}, drawn through the means of the nodes' center distributions in the positions $(p_x,p_y)$, with one color per stage. In each stage panel, the new tree is opaque, the tree drawn by \texttt{SelectRoot} is half faded, the other trees are faded, and circles mark the learned roots. At the third stage, \texttt{SelectRoot} draws the first tree: the third tree restarts at the same node as the second one, which is abandoned.}
    \label{fig:maze}
\end{figure*}

\begin{figure*}[t]
    \centering
    % === ROW 1 ===
    \begin{subfigure}[b]{0.32\textwidth}
        \centering
        \includegraphics[width=\textwidth]{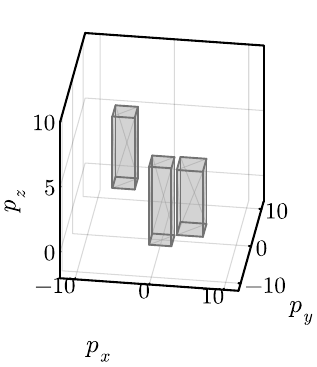}
        \caption{First stage}
        \label{fig:plot1}
    \end{subfigure}
    \hfill
    \begin{subfigure}[b]{0.32\textwidth}
        \centering
        \includegraphics[width=\textwidth]{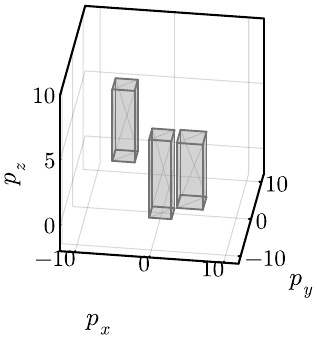}
        \caption{Third stage}
        \label{fig:plot2}
    \end{subfigure}
    \hfill
    \begin{subfigure}[b]{0.32\textwidth}
        \centering
        \includegraphics[width=\textwidth]{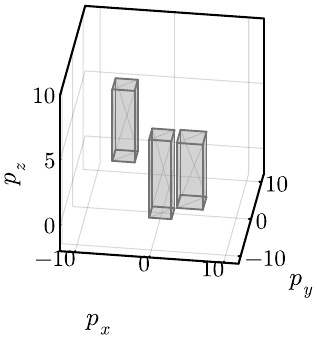}
        \caption{Fourth stage}
        \label{fig:plot3}
    \end{subfigure}
    
    \vspace{0.4cm} % Adjust vertical spacing between rows here
    
    % === ROW 2 ===
    \begin{subfigure}[b]{0.32\textwidth}
        \centering
        \includegraphics[width=\textwidth]{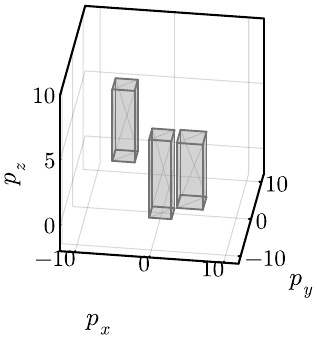}
        \caption{Fifth stage (restarts in tree 3)}
        \label{fig:plot4}
    \end{subfigure}
    \hfill
    \begin{subfigure}[b]{0.32\textwidth}
        \centering
        \includegraphics[width=\textwidth]{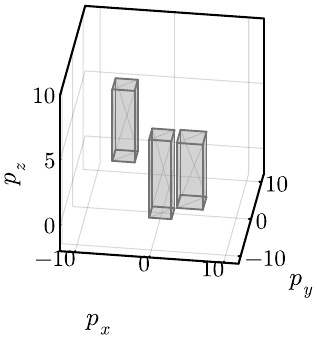}
        \caption{Seventh (last) stage}
        \label{fig:plot5}
    \end{subfigure}
    \hfill
    \begin{subfigure}[b]{0.32\textwidth}
        \centering
        \includegraphics[width=\textwidth]{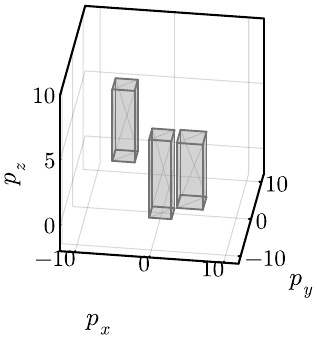}
        \caption{Controlled trajectory}
        \label{fig:plot6}
    \end{subfigure}
    
    \caption{\emph{Double Integrator}, in the positions $(p_x,p_y,p_z)$, with the encoding of Figure~\ref{fig:maze}. The inset of~(a) zooms on the first tree, only two steps deep, in the $(p_x,p_y)$ plane. At the fifth stage, \texttt{SelectRoot} draws the third tree: the fifth tree restarts at the same node as the fourth one, which is abandoned.}
    \label{fig:double_integrator}
\end{figure*}

Table~\ref{tab:experimental_results_comparison_solved} compares three experimental settings: (1) our full Algorithm~\ref{alg:learn-to-propagate-rrt} (\emph{L2P-DR-Forest} in Table~\ref{tab:experimental_results_comparison_solved}), (2) an ablation without uncertainty re-learning (\emph{No re-learning}), i.e. where $N_\text{stage}=1$, and (3) the \emph{$\epsilon$-RandUP-RRT} baseline of \cite{wu2022robust,lew2022randup}, a sampling-based planner that certifies safety by checking that the $\epsilon$-padded per time-step convex hulls of $1{,}000$ randomly propagated particles avoid every obstacle (see Appendix~\ref{section:appendix-randup-rrt} for details). Our key takeaways are:

\textbf{Strong performance on nonlinear systems.} Our approach successfully synthesizes formally safe controllers across challenging environments, achieving a $100\%$ success rate on all five benchmarks, including the most demanding one, \emph{Tractor Trailer}, whose dynamics are given by a learned neural network with a (LipSDP-certified, see Appendix~\ref{section:appendix-tractor-trailer}) Lipschitz constant exceeding one ($1.21$) compounded over the benchmark's $\approx 90$ time steps. Total planning time ranges from $0.08$s on \emph{Switched System} to $55.51$s on \emph{Tractor Trailer}. Importantly, when we use the obtained safe controllers in a Monte-Carlo simulation of the actual underlying system for $5\times 10^3$ simulations, unsafe probabilities are $0\%$ in all time steps for all benchmarks, while target hitting probabilities are $100\%$.

\textbf{Necessity of re-learning.} The uncertainty re-learning mechanism proves essential. The no-relearning ablation reaches $0\%$ success on \emph{all five} benchmarks, despite being granted the same total iteration budget.
Crucially, the MILP optimization required for this stage introduces negligible computational overhead ($\leq 0.12$ s). This is because even with the high confidence considered in our experimental setting, the support size $M$ required by Theorem \ref{thm:learning-ambiguity-ball} to achieve small enough Wasserstein ambiguity set radii remains relatively small, consistently below $10$ for all examples.

\textbf{Comparison against $\epsilon$-RandUP-RRT.} $\epsilon$-RandUP-RRT \cite{wu2022robust,lew2022randup} solves none of the benchmarks more reliably than our method: it reaches $85\%$ on \emph{AV Maze} and $80\%$ on \emph{Unicycle Maze}, matches our $100\%$ on \emph{Double Integrator}, and collapses entirely on the hybrid, mode-switching \emph{Switched System} ($0\%$) and on the highly nonlinear \emph{Tractor Trailer} ($0\%$, timing out on all $20$ seeds). It is also slower whenever it does succeed, markedly so on the mazes (e.g., $31.30$s vs.\ $1.56$s for our method on \emph{AV Maze}), since every extension requires rolling out $1{,}000$ particles through the forward dynamics -- an overhead that grows sharply with the cost of a single forward-dynamics evaluation, as on the neural-network-driven \emph{Tractor Trailer}, whereas our method only ever propagates the much smaller distribution at the center of the ambiguity set. We note that these runs use a $2$-minute per-seed budget; on \emph{Tractor Trailer} every seed reached that budget without terminating, so its $0\%$ should be read as a lower bound on what a longer budget would allow. Crucially, unlike our method, $\epsilon$-RandUP-RRT offers no certified statistical safety guarantee in practice: the confidence bound of \cite{lew2022randup} depends on quantities (e.g., packing numbers of the obstacle boundary) that are difficult to obtain for our benchmarks, so we could not instantiate it at a prescribed confidence level and its results should be read as empirical only.

\textbf{Planning illustration.} Figures~\ref{fig:maze} and~\ref{fig:double_integrator} illustrate the execution of the algorithm on the \emph{AV Maze} and on the \emph{Double Integrator}, depicting the planning trees of the successive stages via the means of the ambiguity sets' center distributions. Each stage roots a new tree, with a freshly learned ambiguity set, at a promising node of a previous tree, so that the trees progressively push the explored region towards the target. On the \emph{AV Maze}, \texttt{SelectRoot} draws the first tree at the third stage: the third tree restarts at the same node as the second one, with a root re-learned from more samples, and the second tree is not used by the returned controller. The fourth \emph{Double Integrator} tree is abandoned in the same way. The first \emph{Double Integrator} tree is only two steps deep (inset of Figure~\ref{fig:double_integrator}(a)), since stage $1$ learns from the fewest samples and its escaped-mass bounds, which accumulate along the tree, are the loosest. Later stages learn from more samples and grow deeper trees. This is the behavior that re-learning enables, whereas a single tree (\emph{No re-learning}) fails on all benchmarks.

\section{Limitations}\label{section:limitations}

A limitation of our approach is its scalability to high-dimensional systems. This limitation stems from the well-known curse of dimensionality affecting Wasserstein ambiguity sets, where the sample complexity required to attain a given approximation accuracy increases rapidly with the dimension of the uncertainty space \cite{fournier2015rate}. We should also stress its dependence on reasonably tight Lipschitz bounds and access to a preferably large dataset of noise samples (or, equivalently, the ability to easily query them). An important future work direction is to investigate alternative probability metrics that alleviate this dependence, such as the sliced Wasserstein distance  \cite{bonneel2015sliced, kolouri2019generalized}. 

\section{Conclusion}\label{section:conclusion}
We introduced Learn-To-Propagate Distributionally-Robust Forest (L2P-DR-Forest), a novel motion-planning algorithm based on distributionally robust optimization that supports nonlinear dynamics and unknown distributions while providing safety and correctness guarantees. By combining optimal transport, statistical learning, and nonlinear uncertainty propagation, L2P-DR-Forest enables the efficient synthesis of safe trajectories for stochastic systems beyond the linear-Gaussian setting. Our experimental results demonstrate that the proposed approach achieves superior performance compared to comparable methods. More broadly, the theoretical framework developed in this work establishes a new connection between Wasserstein-based distribution learning and uncertainty-aware motion planning, opening new directions for scalable and provably safe planning.

%===============================================================================

\bibliographystyle{ACM-Reference-Format}
\bibliography{main}

%===============================================================================

\clearpage
\appendix
This Appendix provides supplementary material organized into five sections. Appendix~\ref{section:appendix-algorithms} introduces the component algorithms underlying Algorithm~\ref{alg:learn-to-propagate-rrt}, followed by Appendix~\ref{section:appendix-proofs}, which contains proofs for all theoretical results presented in this manuscript. Appendix~\ref{section:appendix-heuristics} details the heuristics used by \texttt{SelectRoot} to select a promising node (Algorithm~\ref{alg:select-root}). Appendix~\ref{section:appendix-experiments-details} specifies the environment configurations used in the experiments of Section~\ref{section:experimental-results}, and Appendix~\ref{section:appendix-randup-rrt} details our implementation of the $\epsilon$-RandUP-RRT baseline.

\section{Algorithms}\label{section:appendix-algorithms}

We introduce the algorithms \texttt{LearnAmbiguitySet}, \texttt{SelectRoot}, \texttt{SelectNode}, \texttt{Propagate}, \texttt{ValidityCheck}, and \texttt{TargetCheck}, which are the building blocks of Algorithm \ref{alg:learn-to-propagate-rrt}.

\subsection{Algorithm for \texttt{LearnAmbiguitySet}}
Algorithm \ref{alg:learn-ambiguity} follows naturally from Section \ref{subsection:learning}.

\RestyleAlgo{ruled}
\SetKwComment{Comment}{/* }{ */}
\SetKwProg{Fn}{Function}{:}{}
\SetKwFunction{FLearnAmbiguitySet}{LearnAmbiguitySet}
\begin{algorithm}[ht!]
\DontPrintSemicolon
\caption{Learn high-confidence ambiguity sets from data}
\label{alg:learn-ambiguity}

\Fn{\FLearnAmbiguitySet{$\sD$, $\beta$, $M$}}{

    $\sD_{N_\text{pre}}, \sD_N \gets$ split $\sD$ into disjoint subsets of size $N_\text{pre}, N$

    $B \gets$ bounding box of $\sD_{N_\text{pre}}$ \label{step:build-support}

    $\{ C_i, c_i \}_{i=1}^{M} \gets$ intersections with $\sX$ of the $M$ hyperrectangles obtained from $B$ by cyclic bisections; $c_i\in C_i$ is the center of the $i$-th hyperrectangle if it lies in $\sX$ \label{step:build-partition}

    $p_\ell, p_u \gets $ Clopper-Pearson interval for $\sX \setminus \cup_{i=1}^{M} C_i$ using $\sD_N$ with confidence $\frac{1}{M+1} \beta$

    $\ProbHat, \epsilon(\sD_{N}) \gets $ Theorem \ref{thm:learning-ambiguity-ball} with confidence $\frac{M}{M+1} \beta$

    \Return $\ball_{\epsilon(\sD_{N})}(\ProbHat), (p_\ell, p_u)$
}
\end{algorithm}

The support size $M$ of the discrete approximation $\ProbHat$ is a caller-supplied parameter. Of the dataset's $|\sD|$ samples, $N_\text{pre}$ are allocated to pre-training, leaving $N=|\sD|-N_\text{pre}$ for the training portion $\sD_N$. The pre-training dataset $\sD_{N_\text{pre}}$ defines the bounding box $B=\prod_k[a_k,b_k]$ of its samples, which is partitioned into $M$ hyperrectangles by cyclic bisections. Let $e$ be the dimension of the data and $R\coloneqq\lfloor\log_2M\rfloor$, so that $2^R\leq M<2^{R+1}$. In round $j\in[R]$, every current cell is halved at its midpoint along coordinate $((j-1)\bmod e)+1$, which yields $2^R$ cells. In round $R+1$, only $M-2^R$ of these cells are halved along coordinate $(R\bmod e)+1$, which yields $2^R+(M-2^R)=M$ cells; any choice of the cells to halve is valid as long as it does not depend on $\sD_N$ (e.g., the first ones in the order of construction), and there is no round $R+1$ if $M$ is a power of two. Cells are products of half-open intervals, closed at $b_k$, so that they are disjoint. The cell $C_i$ is the intersection of the $i$-th hyperrectangle with $\sX$, and $c_i\in C_i$ is the center of the hyperrectangle if it lies in $\sX$, and an arbitrary point of $C_i$ otherwise.

\begin{figure*}[t]
    \centering
    \includegraphics[width=0.8\textwidth]{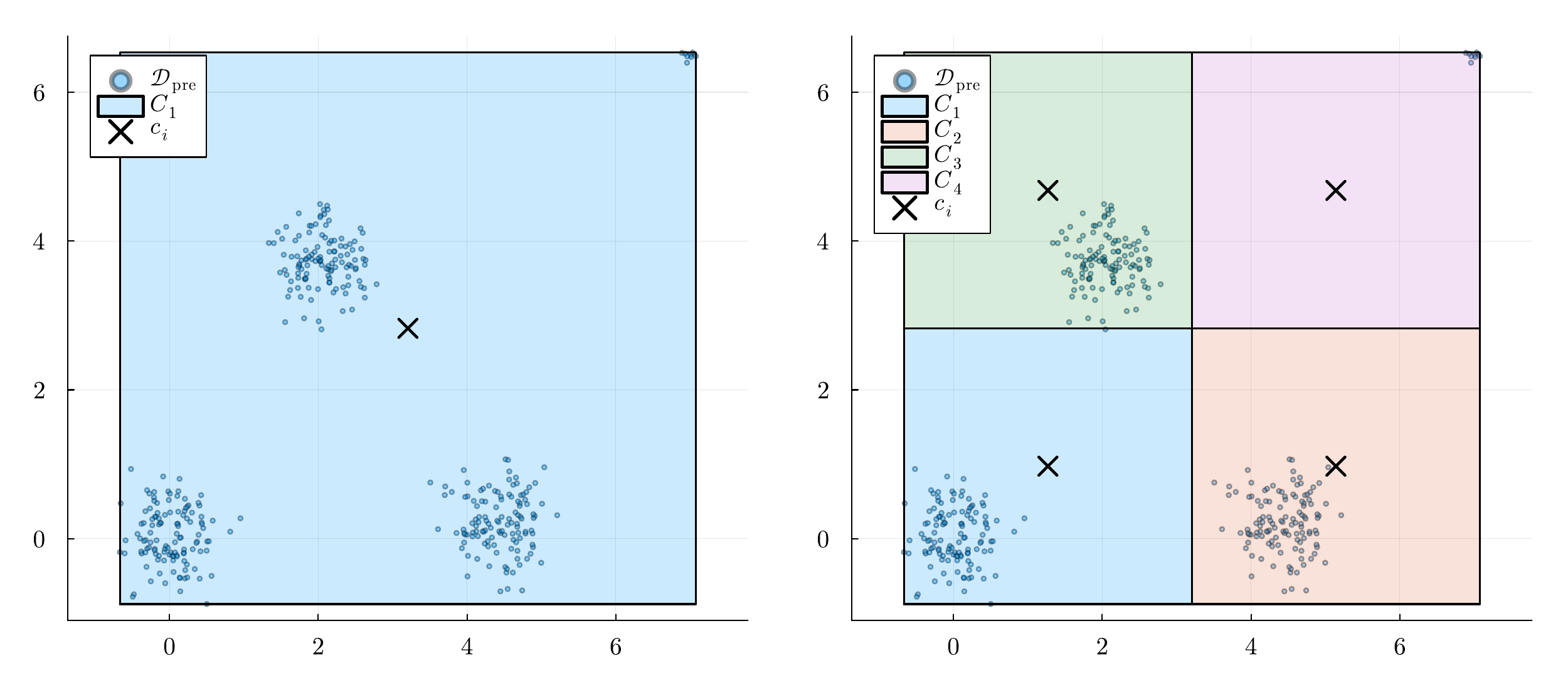}
    \caption{Example of the bounding box of a given artificial dataset $\sD_{\text{pre}}$, partitioned into $M$ cells $C_i$ with centers $c_i$ as explained above, for $M \in \{ 1, 4 \}$.}
    \label{fig:heuristics-partition}
\end{figure*}

\subsection{Algorithm for \texttt{SelectRoot}}
\texttt{SelectRoot} (Algorithm~\ref{alg:select-root}) selects the tree of the forest in which a stage restarts and, in it, the node whose controller roots the tree of the stage. With probability $1-\lambda_1$, it first picks the most recent tree of $\mathbb{F}$, and with probability $\lambda_1$ a tree drawn uniformly from $\mathbb{F}$. Independently, with probability $1-\lambda_2$, it then returns the promising node of that tree, selected by the heuristic of Appendix~\ref{subsection:select-promising-trajectory}, and with probability $\lambda_2$ a node drawn uniformly from that tree. The returned node determines where the new tree is rooted: its own ambiguity set is not reused, but instead learned afresh from new samples at the controller $\texttt{backtrack}(\nu)$ (Lines~\ref{step:roll-forward}--\ref{step:learn-root} of Algorithm~\ref{alg:learn-to-propagate-rrt}), so that the certificate of every refined node is as tight as the sample sizes of the current stage allow. Since the promising node of a tree does not change, drawing an older tree typically restarts at a node where an earlier stage already restarted, with a root re-learned from more samples (Figures~\ref{fig:maze} and~\ref{fig:double_integrator}).

\RestyleAlgo{ruled}
\SetKwComment{Comment}{/* }{ */}
\SetKwProg{Fn}{Function}{:}{}
\SetKwFunction{FSelectRoot}{SelectRoot}
\begin{algorithm}[ht!]
\DontPrintSemicolon
\caption{Select the node at which to root a tree}
\label{alg:select-root}

\Fn{\FSelectRoot{$\mathbb{F}$, $\lambda_1$, $\lambda_2$}}{

    \lIf{$\mathbb{F}=\emptyset$}{\Return $\varnothing$}

    $b_1 \sim \mathrm{Bernoulli}(\lambda_1)$

    \eIf{$b_1=1$}{
        $\texttt{tree}' \gets$ a tree drawn uniformly from $\mathbb{F}$
    }{
        $\texttt{tree}' \gets$ the most recent tree of $\mathbb{F}$
    }

    $b_2 \sim \mathrm{Bernoulli}(\lambda_2)$

    \eIf{$b_2=1$}{
        \Return a node drawn uniformly from $\texttt{tree}'$
    }{
        \Return the promising node of $\texttt{tree}'$ \tcp*[r]{reference tracking (Appendix~\ref{subsection:select-promising-trajectory})}
    }
}
\end{algorithm}

\subsection{Algorithms for \texttt{SelectNode} and \texttt{Propagate}}
In motion planning algorithms, node selection and propagation require an underlying similarity measure  between a state $x \in \sX$ and any node in the tree. In this work, we propose the following pseudometric: letting $\ball_\epsilon(\ProbHat)$ denote the ambiguity set associated with a node $n \in \texttt{tree}$ and $\bar x_n \coloneqq \mathbb E_{y\sim\ProbHat}[y]$ the mean of its center, the distance between node $n$ and $x\in \sX$ is defined as $\mathrm{d}_\text{mean}(\ProbHat, \delta_x) \coloneqq \norm{\bar x_n - x}$. It is a pseudometric between distributions (two distributions with the same mean are at distance zero) that ignores their spread, and it is dominated by the Wasserstein distance between the center and $\delta_x$, $\mathrm{d}_\text{mean}(\ProbHat, \delta_x) \leq \wasserstein_\rho(\ProbHat, \delta_x)$ for every $\rho \geq 1$, by Jensen's inequality (for $\rho=2$ and the Euclidean norm, $\wasserstein_2(\ProbHat, \delta_x)^2 = \mathrm{d}_\text{mean}(\ProbHat, \delta_x)^2 + \operatorname{tr}\mathrm{Cov}(\ProbHat)$, i.e., $\mathrm{d}_\text{mean}$ is $\wasserstein_2$ without its covariance term). Crucially, because the center distributions $\ProbHat$ constructed in this work are always discrete, with at most $M$ atoms, this distance is available in closed form. Ties between nodes at the same distance, e.g., nodes with identical means, are broken uniformly at random.

\RestyleAlgo{ruled}
\SetKwComment{Comment}{/* }{ */}
\SetKwProg{Fn}{Function}{:}{}
\SetKwFunction{FSelectNode}{SelectNode}
\begin{algorithm}[ht!]
\DontPrintSemicolon
\caption{Select a node for further tree expansion}
\label{alg:select-node}

\Fn{\FSelectNode{$\texttt{tree}$}}{

    Uniformly sample a point $x$ in $\sX$ \label{step:sample-x}

    $\texttt{node}_{\text{selected}} \gets$ uniformly random element of $\arg\min_{n \in \texttt{tree}} \mathrm{d}_\text{mean}(\texttt{n.center}, \delta_x)$ \label{step:find-nearest}
    \Return $\texttt{node}_{\text{selected}}, \delta_x$
}
\end{algorithm}

Once the node $\texttt{node}_{\text{selected}}$ closest to $\delta_x$ is identified, we determine the control input $u^\star$ that steers this node as close as possible to $\delta_x$ (for the same distance, among $K$ candidates sampled uniformly from $\sU$), leveraging the theoretical results presented in Section \ref{subsection:propagating}.

\RestyleAlgo{ruled}
\SetKwComment{Comment}{/* }{ */}
\SetKwProg{Fn}{Function}{:}{}
\SetKwFunction{FPropagate}{Propagate}
\begin{algorithm*}[t]
\DontPrintSemicolon
\caption{Propagate (or steer) via ambiguity set propagation}
\label{alg:propagate-node}

\Fn{\FPropagate{$\texttt{node}_{\text{selected}}, \ball_{\epsilon_\vomega}(\ProbHat_\vomega), (p_{\vomega,\ell},p_{\vomega,u}), \delta_x$}}{

    $\ball_{\epsilon_{\text{ref}}}(\ProbHat_{\text{ref}}), (p_\ell^\text{ref},p_u^\text{ref}) \gets \texttt{node}_{\text{selected}}.(\texttt{ambiguity\_set},\texttt{escaped\_mass})$

    Sample $u_1,\dots,u_K$ i.i.d.\ uniformly from $\sU$

    $u^\star \gets \arg\min_{u \in \{u_1,\dots,u_K\}} \mathrm{d}_\text{mean}\big( f_u\# (\ProbHat_{\text{ref}} \otimes \ProbHat_\vomega), \delta_x \big)$ \tcp*[r]{find control for steering} \label{step:before-steer}

    $\ProbHat \gets f_{u^\star} \# (\ProbHat_{\text{ref}} \otimes \ProbHat_\vomega)$ \label{step:steer} \tcp*[r]{propagate center}

    $\ProbHat, \epsilon_{\text{compr}} \gets \texttt{Compress}(\ProbHat, M(s))$ \label{step:compress}

    $\epsilon \gets \sL_{f_{u^\star}} (\epsilon_{\text{ref}}+\epsilon_\vomega) + \epsilon_{\text{compr}}$ \label{step:propagate-radius} \tcp*[r]{propagate ambiguity radius}

    $p_\ell \gets p_\ell^\text{ref} + p_{\vomega,\ell} - p_\ell^\text{ref} p_{\vomega,\ell}$, \quad $p_u \gets p_u^\text{ref} + p_{\vomega,u} - p_u^\text{ref} p_{\vomega,u}$ \label{step:propagate-escaped-mass} \tcp*[r]{propagate escaped mass}

    \Return $\big( \ball_\epsilon(\ProbHat), (p_\ell,p_u), u^\star \big)$
}
\end{algorithm*}
Line~\ref{step:before-steer} of Algorithm~\ref{alg:propagate-node} approximates the $\arg\min$ over $\sU$ by its minimizer over $K$ candidates sampled uniformly from $\sU$. The candidates are drawn independently of the node and of $\delta_x$. The distance in line~\ref{step:before-steer} only requires the mean of the propagated distribution, which is available in closed form from its atoms. Lines~\ref{step:steer} and \ref{step:propagate-radius} are extensively discussed in Section \ref{subsection:propagating}, and line~\ref{step:propagate-escaped-mass} (propagating the escaped mass) in Proposition~\ref{prop:specialize-prop-to-problem} below. Over multiple iterations, the support size of the resulting discrete distribution $\ProbHat$ in line \ref{step:steer} may grow, thereby increasing computational and memory overhead. To mitigate this issue, we introduce a compression step in line \ref{step:compress}. This step compresses $\ProbHat$ back to at most $M(s)$ atoms, the support size of the stage's root: it partitions the bounding box of the atoms of $\ProbHat$ into $M(s)$ cells by cyclic bisections, as in \texttt{LearnAmbiguitySet}, and merges the atoms of each cell into a single atom located at their probability-weighted mean, which lies in the cell. The Wasserstein distance between $\ProbHat$ and its compressed counterpart is at most the cost $\epsilon_\text{compr}$ of the coupling that moves every atom to the mean of its cell.

\subsection{Algorithms for \texttt{ValidityCheck} and \texttt{TargetCheck}}
Before introducing the algorithms for \texttt{ValidityCheck} and \texttt{TargetCheck}, we introduce Proposition \ref{prop:specialize-prop-to-problem} which extends and specializes Proposition~\ref{prop:extend-to-non-compact} to Problem \ref{problem:reach-avoid}. In the following, let $f_u = f(\cdot, u, \cdot)$ for any given $u \in \sU$, and $\bm{u} \in \sU^{k}$ a given controller.
\begin{proposition}\label{prop:specialize-prop-to-problem}
    Let the state distribution induced by $\bm{u}$, $\Prob^{\bm{u}}_{x_k}$, be such that $\Prob^{\bm{u}}_{x_k}|_{\sX_c} \in \ball_{\epsilon_x}(\ProbHat_{x})$ and $\Prob^{\bm{u}}_{x_k}(\sX \setminus \sX_c) \in [p_\ell, p_u]$ for some bounded set $\sX_c \subset \sX$. Analogously, let the process noise distribution, $\Prob_\vomega$, be such that $\Prob_\vomega|_{\sW_c} \in \ball_{\epsilon_\vomega}(\ProbHat_{\vomega})$ and $\Prob_\vomega(\sW \setminus \sW_c) \in [p_{\vomega,\ell}, p_{\vomega,u}]$ for some bounded set $\sW_c \subset \sW$. Then, for any set $\sS \in \sB(\sX)$,
    \begin{multline}
        f_u \# (\Prob^{\bm{u}}_{x_k} \otimes \Prob_\vomega)(\sS) \leq \\
        (1-p_\ell)(1-p_{\vomega,\ell}) \cdot \sup_{\ProbQ \in \ball_{\epsilon}(f_u \# (\ProbHat_{x} \otimes \ProbHat_{\vomega}))} \ProbQ(\sS) + \big( p_u + p_{\vomega,u} - p_u p_{\vomega,u}\big),
    \end{multline}
    where $\epsilon=\sL_{f_u}(\epsilon_x + \epsilon_\vomega)$.
\end{proposition}
Because Proposition~\ref{prop:specialize-prop-to-problem} bounds the escaped mass after a \emph{single} application of $f_u$, it is applied once per \texttt{Propagate}() call, using the \emph{immediate parent} node's escaped-mass bound $(p_\ell^\text{ref},p_u^\text{ref})$, as in line~\ref{step:propagate-escaped-mass} of Algorithm~\ref{alg:propagate-node}.
Both checks take a node as input. For a node storing $\ball_\epsilon(\ProbHat)$ and $(p_\ell,p_u)$, define
\begin{equation}\label{eq:node-bound}
    \pi(\sS) \coloneqq (1-p_\ell) \sup_{\ProbQ \in \ball_{\epsilon}(\ProbHat)} \ProbQ(\sS) + p_u, \qquad \sS \in \sB(\sX).
\end{equation}
Propositions~\ref{prop:extend-to-non-compact} and~\ref{prop:specialize-prop-to-problem} show that $\pi(\sS)$ upper-bounds the probability of $\sS$ under the state distribution $\Prob$ induced by the controller of the node, with high confidence. \texttt{ValidityCheck} (Algorithm~\ref{alg:validity-check}) therefore returns whether $\pi(\sS)\leq\varepsilon_\sS$, which then certifies $\Prob(\sS)\leq\varepsilon_\sS$. Since $\Prob(\sT)\geq 1-\varepsilon_\sT$ is equivalent to $\Prob(\sX\setminus\sT)\leq\varepsilon_\sT$, \texttt{TargetCheck}$(\texttt{node},\sT,\varepsilon_\sT)$ is analogous and returns \texttt{ValidityCheck}$(\texttt{node},\sX\setminus\sT,\varepsilon_\sT)$.

\RestyleAlgo{ruled}
\SetKwProg{Fn}{Function}{:}{}
\SetKwFunction{FValidityCheck}{ValidityCheck}
\begin{algorithm}[ht!]
\DontPrintSemicolon
\caption{Check whether a node certifies $\Prob(\sS)\leq\varepsilon_\sS$}
\label{alg:validity-check}

\Fn{\FValidityCheck{$\texttt{node}$, $\sS$, $\varepsilon_\sS$}}{

    $\ball_\epsilon(\ProbHat), (p_\ell,p_u) \gets \texttt{node}.(\texttt{ambiguity\_set},\texttt{escaped\_mass})$

    $q \gets \texttt{SolveSup}\big(\ball_\epsilon(\ProbHat), \sS\big)$ \tcp*[r]{Algorithm~\ref{alg:compute-bounds-in-probability}}

    \Return $(1-p_\ell)\,q + p_u \leq \varepsilon_\sS$
}
\end{algorithm}

\begin{figure}[h]
    \centering
    \includegraphics[width=\linewidth]{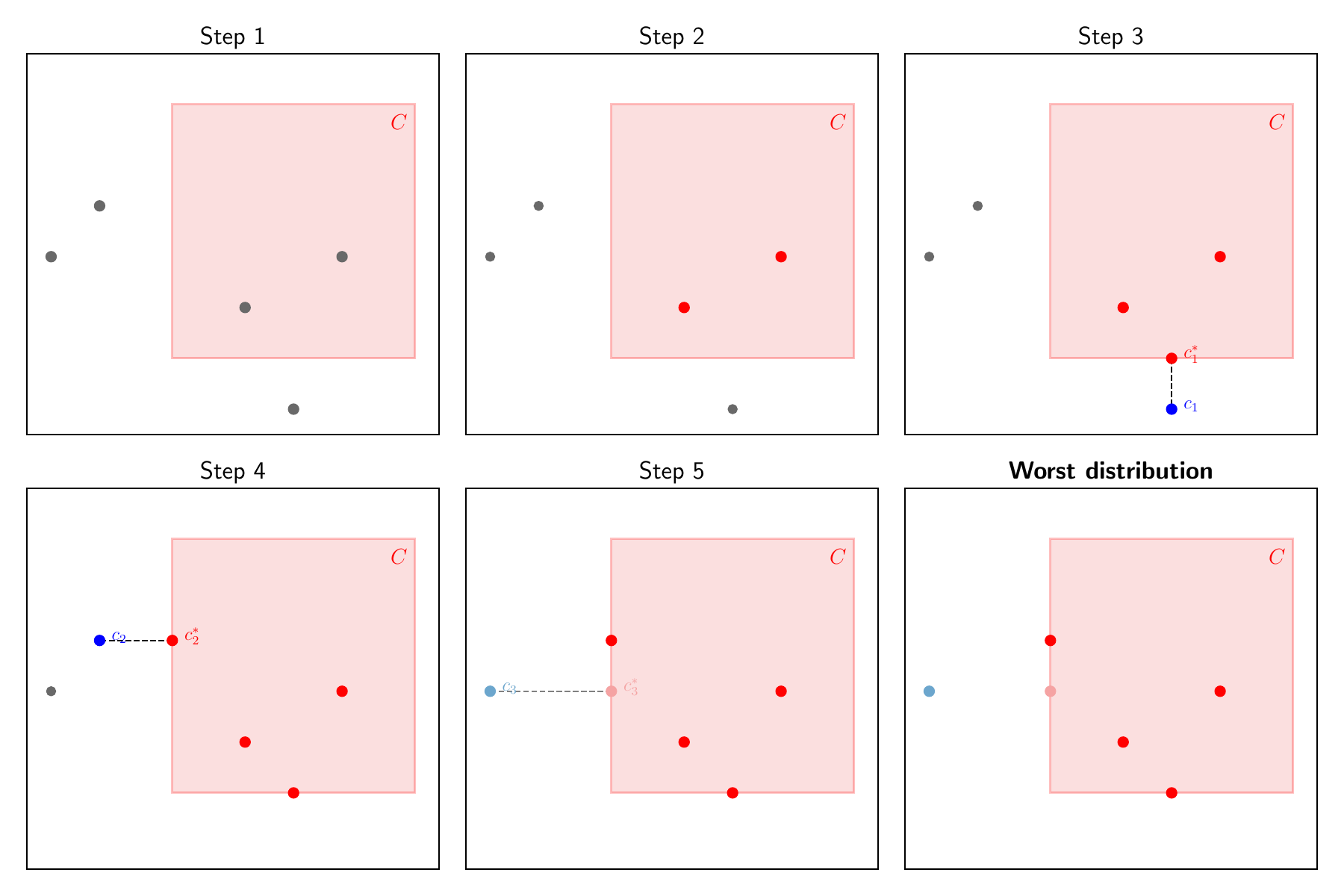}
    \caption{Representation of the worst-case probability mass displacement given by Algorithm \ref{alg:compute-bounds-in-probability}.}
    \label{fig:dro-algo}
\end{figure}

To conclude, it remains to show how to efficiently solve $\sup_{ \ProbQ \in \ball_{\epsilon}(\ProbHat) } \ProbQ(\sS)$ for any set $\sS \in \sB(\sX)$. For this, we introduce \texttt{SolveSup} (Algorithm \ref{alg:compute-bounds-in-probability}), which solves this problem by following Example~7 in \cite{gao2023distributionally}. In line 2, we compute the distance from each support point of the discrete distribution $\ProbHat$ to the set $\sS$. For the points already in $\sS$, we sum their probabilities (line 4). For the outside points, we sort them ascendingly by distance to $\sS$ (line 6). Lines 7-16 iteratively move as much probability mass to the set $\sS$ as allowed by the Wasserstein budget $\epsilon$, starting from closer locations. Figure \ref{fig:dro-algo} depicts a schematic representation of Algorithm \ref{alg:compute-bounds-in-probability}.
\RestyleAlgo{ruled}
\SetKwComment{Comment}{/* }{ */}
\begin{algorithm}[h]
\DontPrintSemicolon
\caption{Solving $\sup_{\ProbQ\in \ball_{\epsilon}(\ProbHat)} \ProbQ(\sS)$ for a discrete distribution $\ProbHat$ and a set $\sS \in \sB(\sX)$ (Example 7 in \cite{gao2023distributionally}).}\label{alg:compute-bounds-in-probability}
\KwIn{Ambiguity set $\ball_{\epsilon}(\ProbHat)$, where $\ProbHat=\sum_{i=1}^{M} \evpi{i}\delta_{c_i}$, and set $\sS\in\sB(\sX)$}
\KwOut{Result of $\sup_{\ProbQ\in \ball_{\epsilon}(\ProbHat)} \ProbQ(\sS)$}
\SetKwFunction{FMain}{SolveSup}
\SetKwProg{Fn}{function}{:}{}
\Fn{\FMain{$\ball_{\epsilon}(\ProbHat), \sS$}}{
    $d \gets \big[ \inf_{z\in \sS} \norm{z - c_{i}}^\rho \text{ for } i \in [M] \big]$
    
    $I \gets \big\{ i \text{ s.t. } d^{(i)}>0 \big\}$, $I_0 \gets \big\{i \text{ s.t. } d^{(i)}=0\big\}$
    
    $p \gets \sum_{i\in I_0} \evpi{i}$
    
    $\wasserstein \gets \epsilon^\rho$
    
    $I \gets \texttt{sort ascendingly according to }d$
    
    \For{$i \in I$}{
        \If{$\wasserstein = 0$}{\texttt{break}}
        \Else{
            \If{$\evpi{i} \times d^{(i)} \leq \wasserstein$}{
            $p \gets p + \evpi{i}$\;
            $\wasserstein \gets \wasserstein - \evpi{i} \times d^{(i)}$
        }\Else{
            $p \gets p + \frac{\wasserstein}{d^{(i)}}$\;
            \texttt{break}
        }
        }
    }
    \Return{$p$}
}
\end{algorithm}

\section{Proofs}\label{section:appendix-proofs}

\subsection{Proof of Theorem \ref{thm:learning-ambiguity-ball}}
\begin{proof}
    The proof follows directly from Theorem 3.1 in \cite{figueiredo2026efficient} after noting that gathering the $N_{\sX_c} \leq N$ samples from $\sD_N$ that lie on the set $\sX_c$ is equivalent to constructing a dataset $\sD_{N_{\sX_c}}$ with $N_{\sX_c}$ i.i.d. samples from the distribution $\Prob|_{\sX_c}$.
\end{proof}

\subsection{Proof of Proposition \ref{prop:extend-to-non-compact}}
\begin{proof}
    By the law of total probability, we have that for any set $\sS \subseteq \sX$,
    \begin{equation}\label{eq:proof-aux-eq-8}
        \Prob(\sS) = \Prob(\sS \cap \sX_c) + \Prob(\sS \cap \sX \setminus \sX_c).
    \end{equation}
    We bound the two terms separately. We start with the second term. Since $\sS \cap (\sX \setminus \sX_c) \subseteq \sX \setminus \sX_c$, we have $\Prob(\sS \cap (\sX \setminus \sX_c)) \leq \Prob(\sX \setminus \sX_c)$. We then proceed to bound the first term. By definition of the normalized restriction $\Prob|_{\sX_c}$, we have that

    \begin{equation}\label{eq:proof-aux-eq-9}
        \Prob(\sS \cap \sX_c) = \Prob|_{\sX_c}(\sS) \cdot \Prob(\sX_c) \leq \Prob|_{\sX_c}(\sS) \cdot (1-\Prob(\sX \setminus \sX_c)).
    \end{equation}
    Substituting those bounds in \eqref{eq:proof-aux-eq-8}, and using the fact that, by Clopper-Pearson Lemma, it holds with probability $1-\tilde\beta$ over $\sD_N$ that $p_\ell \leq \Prob(\sX \setminus \sX_c) \leq p_u$, we have that
    \begin{equation}
        \Prob(\sS) \leq \Prob|_{\sX_c}(\sS) \cdot (1-p_\ell) + p_u.
    \end{equation}
    Finally, we note that, from Theorem~\ref{thm:learning-ambiguity-ball}, with confidence $1-\beta$, $\Prob|_{\sX_c} \in \ball_\epsilon(\ProbHat)$. Therefore, by applying the union bound, with probability at least $1-\beta-\tilde\beta$ over $\sD_N$ it must hold that
    \begin{equation}
        \Prob(\sS) \leq (1-p_\ell) \cdot \sup_{\ProbQ \in \ball_\epsilon(\ProbHat)} \ProbQ(\sS) + p_u,
    \end{equation}
    which concludes the proof.
\end{proof}

\subsection{Proof of Proposition~\ref{prop:specialize-prop-to-problem}}\label{subsection:escaped-ambiguity}
Soundness (Theorem~\ref{thm:correctness-algorithm-learn-to-propagate}) and completeness (Theorem~\ref{thm:probabilistic-completeness}) rest on one invariant: the true state distribution of every node belongs to the ambiguity set defined by the tuple $(\ProbHat,\epsilon,p_\ell,p_u)$ stored at that node (Definition~\ref{def:escaped-ambiguity}). Throughout, $\sL_f$ is the joint Lipschitz constant of $f$, i.e., $\norm{f(x,u,\vomega)-f(x',u',\vomega')}\leq\sL_f\big(\norm{x-x'}+\norm{u-u'}+\norm{\vomega-\vomega'}\big)$, so that $f_u\coloneqq f(\cdot,u,\cdot)$ is $\sL_{f_u}$-Lipschitz on $\sX\times\sW$, equipped with the norm $\norm{x}+\norm{\vomega}$, with the constant $\sL_{f_u}\leq\sL_f$ used by Algorithm~\ref{alg:propagate-node}. Also, $a\oplus b\coloneqq a+b-ab=1-(1-a)(1-b)$ for $a,b\in[0,1]$, which is nondecreasing in each argument.

\begin{definition}[Ambiguity set with escaped mass]\label{def:escaped-ambiguity}
For a discrete distribution $\ProbHat$, a radius $\epsilon\geq0$ and $0\leq p_\ell\leq p_u\leq1$, let
\begin{multline*}
\ambesc(\ProbHat,\epsilon,p_\ell,p_u)\coloneqq\big\{(1-q)\mu+q\nu:\ \mu\in\ball_\epsilon(\ProbHat),\ \\ \nu\text{ a probability measure},\ q\in[p_\ell,p_u]\big\},
\end{multline*}
i.e., a Wasserstein ball around $\ProbHat$ for the mass $1-q$ that has not escaped, plus an arbitrary distribution $\nu$ for the escaped mass $q$. For $\sS\in\sB(\sX)$, $\pi(\sS)$ denotes the bound~\eqref{eq:node-bound} of this tuple, which is tested by \texttt{ValidityCheck} and \texttt{TargetCheck}.
\end{definition}

\begin{lemma}[Calculus of ambiguity sets with escaped mass]\label{lem:escaped-ambiguity}
It holds that:

\emph{(a) Validity.} If $\Prob|_{\sX_c}\in\ball_\epsilon(\ProbHat)$ and $\Prob(\sX\setminus\sX_c)\in[p_\ell,p_u]$, then $\Prob\in\ambesc(\ProbHat,\epsilon,p_\ell,p_u)$. 

\emph{(b) Upper-bound.} Further, every $\Prob\in\ambesc(\ProbHat,\epsilon,p_\ell,p_u)$ satisfies $\Prob(\sS)\leq\pi(\sS)$ for all $\sS\in\sB(\sX)$.

\emph{(c) Propagation.}\footnote{Algorithm~\ref{alg:propagate-node} also compresses the center distribution (line~\ref{step:compress}). If $\ProbHat'$ satisfies $\wasserstein_\rho\big(f_u\#(\ProbHat_x\otimes\ProbHat_\vomega),\ProbHat'\big)\leq\epsilon_\text{c}$, the triangle inequality shows that the statement trivially holds with center $\ProbHat'$ and radius $\sL_{f_u}(\epsilon_x+\epsilon_\vomega)+\epsilon_\text{c}$.} Let $\Prob_x\in\ambesc(\ProbHat_x,\epsilon_x,p_\ell,p_u)$ and $\Prob_\vomega\in\ambesc(\ProbHat_\vomega,\epsilon_\vomega,p_{\vomega,\ell},p_{\vomega,u})$, and let $u\in\sU$. Then
\begin{multline*}
f_u\#(\Prob_x\otimes\Prob_\vomega)\in\ambesc\big(f_u\#(\ProbHat_x\otimes\ProbHat_\vomega),\ \sL_{f_u}(\epsilon_x+\epsilon_\vomega),\ \\ p_\ell\oplus p_{\vomega,\ell},\ p_u\oplus p_{\vomega,u}\big).
\end{multline*}
\end{lemma}
\begin{proof}
(a) Take $\mu=\Prob|_{\sX_c}$, $\nu=\Prob|_{\sX\setminus\sX_c}$, $q=\Prob(\sX\setminus\sX_c)$, which shows the first claim. 

(b) If $\Prob\in\ambesc(\ProbHat,\epsilon,p_\ell,p_u)$, then $\Prob(\sS)=(1-q)\mu(\sS)+q\nu(\sS)\leq(1-p_\ell)\sup_{\ball_\epsilon(\ProbHat)}\ProbG(\sS)+p_u$, as $1-q\leq1-p_\ell$ and $q\nu(\sS)\leq q\leq p_u$.

(c) Write $\Prob_x=(1-q)\mu+q\nu$ and $\Prob_\vomega=(1-q_\vomega)\mu_\vomega+q_\vomega\nu_\vomega$ as in the definition. Expanding the product,
$\Prob_x\otimes\Prob_\vomega=(1-q')\,\mu\otimes\mu_\vomega+q'\nu'$, where $q'\coloneqq q\oplus q_\vomega\in[p_\ell\oplus p_{\vomega,\ell},\,p_u\oplus p_{\vomega,u}]$ and $\nu'\coloneqq\big(q\,\nu\otimes\Prob_\vomega+(1-q)q_\vomega\,\mu\otimes\nu_\vomega\big)/q'$ (arbitrary if $q'=0$). Hence $f_u\#(\Prob_x\otimes\Prob_\vomega)=(1-q')f_u\#(\mu\otimes\mu_\vomega)+q'f_u\#\nu'$, and it remains to show that $f_u\#(\mu\otimes\mu_\vomega)$ lies in the claimed ball around $f_u\#(\ProbHat_x\otimes\ProbHat_\vomega)$. Indeed,  applying Minkowski's inequality gives $\wasserstein_\rho(\mu\otimes\mu_\vomega,\ProbHat_x\otimes\ProbHat_\vomega)\leq\epsilon_x+\epsilon_\vomega$. Finally, pushing a coupling through $(f_u,f_u)$ gives $\wasserstein_\rho(f_u\#\alpha,f_u\#\alpha')\leq\sL_{f_u}\wasserstein_\rho(\alpha,\alpha')$.
\end{proof}

\begin{proof}[Proof of Proposition~\ref{prop:specialize-prop-to-problem}]
Given the assumptions on $\Prob^{\bm u}_{x_k}$ and $\Prob_\vomega$, it comes trivially from Lemma~\ref{lem:escaped-ambiguity}(a) that $\Prob^{\bm u}_{x_k}\in\ambesc(\ProbHat_x,\epsilon_x,p_\ell,p_u)$ and $\Prob_\vomega\in\ambesc(\ProbHat_\vomega,\epsilon_\vomega,p_{\vomega,\ell},p_{\vomega,u})$. Then, from Lemma~\ref{lem:escaped-ambiguity}(c), we have that $f_u\#(\Prob^{\bm u}_{x_k}\otimes\Prob_\vomega)\in\ambesc\big(f_u\#(\ProbHat_x\otimes\ProbHat_\vomega),\epsilon,p_\ell\oplus p_{\vomega,\ell},p_u\oplus p_{\vomega,u}\big)$, with $\epsilon=\sL_{f_u}(\epsilon_x+\epsilon_\vomega)$. Finally, applying Lemma~\ref{lem:escaped-ambiguity}(b) concludes the proof, since $1-p_\ell\oplus p_{\vomega,\ell}=(1-p_\ell)(1-p_{\vomega,\ell})$.
\end{proof}

\subsection{Proof of Theorem~\ref{thm:correctness-algorithm-learn-to-propagate}}
\begin{proof}
Let $\sF_s$ be the $\sigma$-algebra generated by all data consumed and all internal random draws made up to the end of stage $s$, together with the random draws of \texttt{SelectRoot} at stage $s+1$ ($\sF_0$ is trivial, up to those of stage $1$).

\emph{Step 1: all learning calls succeed with probability at least $1-\beta$.} Let $\sE$ be the event that every call of \texttt{LearnAmbiguitySet} returns a tuple whose ambiguity set contains its target distribution (Lemma~\ref{lem:escaped-ambiguity}(a)). Stage $s$ makes two calls, with targets $Q_s\coloneqq\Prob^{\bm u}_{x_k}$ (root, line~\ref{step:learn-root}) and $\Prob_\vomega$ (noise, line~\ref{step:learn-noise}), where $\bm u=\texttt{backtrack}(\nu)$ for the node $\nu$ returned by \texttt{SelectRoot} at stage $s$, and $k=|\bm u|$. The node $\nu$ is a function of the forest, which is built by the stages before $s$, and of the draws of \texttt{SelectRoot} at stage $s$ (the reference trajectory of the heuristic of Appendix~\ref{subsection:select-promising-trajectory} is computed beforehand from samples independent of $\sD_{x_0},\sD_\vomega$), and is therefore $\sF_{s-1}$-measurable. Stage $s$ consumes not-yet-used, disjoint blocks of $\sD_{x_0}$ and $\sD_\vomega$ whose positions are determined by $\sF_{s-1}$. Since both datasets are i.i.d., these blocks are independent of $\sF_{s-1}$. Rolling the blocks $\sD_{x_0}^{(s)},\sD_\vomega^{(s)}$ forward under $\bm u$ (line~\ref{step:roll-forward}) yields i.i.d.\ samples from $Q_s$, and the block $\sD_{\vomega,N_\vomega(s)}^{(s)}$ consists of i.i.d.\ samples from $\Prob_\vomega$. Conditionally on $\sF_{s-1}$, both are thus \emph{fixed} distributions, and Theorem~\ref{thm:learning-ambiguity-ball} and Proposition~\ref{prop:extend-to-non-compact} (with the internal split of the confidence level in Algorithm~\ref{alg:learn-ambiguity}) give failure probabilities at most $\zeta\beta_s$ and $(1-\zeta)\beta_s$ for the two calls, hence at most $\beta_s$ for stage $s$ given $\sF_{s-1}$, and unconditionally. Boole's inequality gives $\Prob(\sE^c)\leq\sum_{s}\beta_s\leq\beta$.

\emph{Step 2: soundness on $\sE$.} Fix an outcome in $\sE$. By induction on the order of creation, every node $n$ ever created stores a tuple whose ambiguity set contains $\Prob^{\bm u_n}_{x_{k_n}}$, where $\bm u_n\coloneqq\texttt{backtrack}(n)$ and $k_n\coloneqq|\bm u_n|$: for a root this is $\sE$, and for a child it is Lemma~\ref{lem:escaped-ambiguity}(c) applied to the parent's tuple and the noise tuple of the same stage, for whichever control $u^\star$ was used, followed by the compression of the center, which adds its radius $\epsilon_\text{c}$ to the radius by the triangle inequality (footnote of Lemma~\ref{lem:escaped-ambiguity}). This is exactly what lines~\ref{step:compress}, \ref{step:propagate-radius} and~\ref{step:propagate-escaped-mass} of Algorithm~\ref{alg:propagate-node} compute. By Lemma~\ref{lem:escaped-ambiguity}(b), $\Prob^{\bm u_n}_{x_{k_n}}(\sS)\leq\pi_n(\sS)$ for every $\sS$, and $\pi_n(\sO)$, $\pi_n(\sX\setminus\sT)$ are what \texttt{ValidityCheck} and \texttt{TargetCheck} compare with $\varepsilon_\sO,\varepsilon_\sT$. This holds for every node, accepted or not.

\emph{Step 3: conclusion.} Suppose the algorithm returns $\bm u_{H-1}$ on an outcome in $\sE$. Each $u_{k-1}$, $k\in[H]$, labels a node with prefix $\bm u_{k-1}$ that was accepted by \texttt{ValidityCheck} when created (by induction over the stages: the controller stored in a root is empty or the backtrack of a node of an earlier tree, whose prefixes are thus backtracks of accepted nodes, and the nodes of a tree other than its root were accepted when created), so $\Prob^{\bm u_{k-1}}_{x_k}(\sO)\leq\pi(\sO)\leq\varepsilon_\sO$ by Step~2, while the last node passed \texttt{TargetCheck}, so $\Prob^{\bm u_{H-1}}_{x_H}(\sX\setminus\sT)\leq\varepsilon_\sT$. Both constraints of Problem~\ref{problem:reach-avoid} thus hold on $\sE$, i.e., with probability at least $1-\beta$.

The argument is unchanged if $N_\text{stage}=\infty$: Boole's inequality extends to countably many calls whose confidence levels sum to at most $\beta$.
\end{proof}

\subsection{Proof of Theorem~\ref{thm:probabilistic-completeness}}\label{subsection:proof-completeness}
\paragraph{Auxiliary result $\#1$: transferring check.}
\texttt{ValidityCheck} and \texttt{TargetCheck} test the bound $\pi$ of the tuple $(\ProbHat,\epsilon,p_\ell,p_u)$ stored at a node, whereas Assumption~\ref{assumption:robust-feasibility} gives a margin only for the worst-case probability over $\Delta$-balls around the nominal state distributions induced by $\tilde{\bm u}$. The following lemma transfers this margin to the checks: if the true distribution $\Prob$ of the node is within $\Delta-2\epsilon$ of a reference distribution $\ProbQ$ (later, a nominal one), then $\pi(\sS)$ exceeds the worst-case probability of $\sS$ over $\ball_\Delta(\ProbQ)$ by at most $2p_u$, which is small for tight tuples.
\begin{lemma}[Threshold transfer]\label{lem:transfer}
Let $\Prob\in\ambesc(\ProbHat,\epsilon,p_\ell,p_u)$ and $\ProbQ\in\sP_\rho(\sX)$ be such that $\wasserstein_\rho(\Prob,\ProbQ)\leq\Delta-2\epsilon$. Then, for all $\sS\in\sB(\sX)$,
\[
\pi(\sS)\ \leq\ \sup_{\ProbG\in\ball_\Delta(\ProbQ)}\ProbG(\sS) + 2p_u .
\]
\end{lemma}
\begin{proof}
Using Definition~\ref{def:escaped-ambiguity}, $\Prob\in\ambesc(\ProbHat,\epsilon,p_\ell,p_u)$ means that there exists $\mu\in\ball_\epsilon(\ProbHat)$ and $q\in[p_\ell,p_u]$ such that $\Prob=(1-q)\mu+q\nu$. Now, let $\ProbG\in\ball_\epsilon(\ProbHat)$ be any arbitrary measure, and define $\ProbG'=(1-q)\ProbG+q\nu$. From the joint convexity of $\wasserstein_\rho^\rho$, we have that $\wasserstein_\rho^\rho(\ProbG',\Prob)\leq (1-q)\wasserstein_\rho^\rho(\ProbG,\mu)+q\wasserstein_\rho^\rho(\nu,\nu)\leq (2\epsilon)^\rho$, where we use that $\wasserstein_\rho(\ProbG,\mu) \leq 2\epsilon$ as both distributions belong to the $\wasserstein_\rho$-ball of radius $\epsilon$ around $\ProbHat$, and $\wasserstein_\rho(\nu,\nu)=0$. Hence $\wasserstein_\rho(\ProbG',\Prob)\leq 2\epsilon$. From the triangle inequality, $\wasserstein_\rho(\ProbG',\ProbQ)\leq \wasserstein_\rho(\ProbG',\Prob)+\wasserstein_\rho(\Prob,\ProbQ)\leq 2\epsilon + (\Delta-2\epsilon) = \Delta$. Moreover, we observe that by construction $\ProbG'(\sS) \geq (1-q)\ProbG(\sS)$ and $0 \leq q-p_\ell \leq p_u$. Thus,
\begin{align*}
    (1-p_\ell)\ProbG(\sS)&=(1-q)\ProbG(\sS) + (q-p_\ell)\ProbG(\sS)\leq (1-q)\ProbG(\sS) + p_u \\
    &\leq \ProbG'(\sS) + p_u \leq \sup_{\tilde\ProbG\in\ball_\Delta(\ProbQ)}\tilde\ProbG(\sS) + p_u,
\end{align*}
since $\ProbG' \in \ball_\Delta(\ProbQ)$. Adding $p_u$ to both sides and taking the supremum over $\ProbG$ in $\ball_\epsilon(\ProbHat)$ on the left side gives the claim.
\end{proof}

\paragraph{Auxiliary result $\#2$: learner consistency.}
Now, we move to a second auxiliary result and we prove that our distribution learner is \emph{consistent}: under some scheduling conditions, both the learned radius $\epsilon$ and the escaped-mass upper bound $p_u$ can be controlled to any desired level.

In the following, let $\kappa\geq1$ be such that $\norm{x}\leq\kappa\norm{x}_\infty$ on $\realNum^d$ and $\realNum^w$. Partition a box with longest side $h$ in $\realNum^e$ into $M$ cells by cyclic bisections (Appendix~\ref{section:appendix-algorithms}), and let $R=\lfloor\log_2M\rfloor$. Every cell has been obtained by at least $R$ cyclic bisections, so every coordinate has been halved at least $\lfloor R/e\rfloor=\lfloor\log_2M/e\rfloor>\log_2M/e-1$ times. The $M$ cells thus have longest side at most $h2^{-\lfloor\log_2M/e\rfloor}\leq2hM^{-1/e}$, hence diameter at most
\begin{equation}\label{eq:bisection}
    2\kappa h\,M^{-1/e}.
\end{equation}

\begin{lemma}[Consistency]\label{lem:consistency}
Let $\xi_1,\dots,\xi_{n+N}$ be i.i.d.\ samples of a $\sigma$-sub-Gaussian distribution on $\realNum^e$, and consider the call $\texttt{LearnAmbiguitySet}((\xi_i)_i,\beta',M)$ with pre-training samples $\xi_1,\dots,\xi_n$ and training samples $\xi_{n+1},\dots,\xi_{n+N}$. Let $\tau\in(0,1]$, $\Lambda\coloneqq\ln(2(M+1)/\beta')$, $t\coloneqq\sqrt{\Lambda/N}$ and $r\coloneqq2\sigma\sqrt{\ln(2(n+N)^2)}$. With probability at least $1-4e/(\tau(n+1))-1/(n+N)$, the returned radius $\epsilon$ and escaped-mass upper bound $p_u$ satisfy
\[
p_u\leq\tfrac\tau2+t,\qquad\epsilon\leq\kappa r\big(2M^{-1/e}+(Mt/2)^{1/\rho}\big).
\]
\end{lemma}
\begin{proof}
Let $B$ be the bounding box of the pre-training samples and $h$ its longest side. By the union bound and the sub-Gaussian tail, with probability at least $1-n\cdot2\exp(-\ln(2(n+N)^2))\geq1-1/(n+N)$ all pre-training samples lie in the cube of center $m$ and half-side $\sigma\sqrt{\ln(2(n+N)^2)}$, so that $h\leq r$.

Let $N_\text{out}$ be the number of training samples outside $B$. In each coordinate, at most one of $n+1$ i.i.d.\ values is strictly smaller (larger) than all the others, so by exchangeability a fresh sample lies outside $B$ with probability at most $2e/(n+1)$. Hence $\expect[N_\text{out}/N]\leq2e/(n+1)$, and $N_\text{out}/N\leq\tau/2$ with probability at least $1-4e/(\tau(n+1))$ by Markov's inequality. We work on this event and on the event above. On the former event, $N_{\sX_c}=N-N_\text{out}\geq N/2$.

By Algorithm~\ref{alg:learn-ambiguity}, Theorem~\ref{thm:learning-ambiguity-ball} and Proposition~\ref{prop:extend-to-non-compact}, the bounds computed by the call are one-sided Clopper--Pearson bounds of level $\beta'/(2(M+1))=e^{-\Lambda}$. Let $\bar p$ be such an upper bound for $Z$ successes in $n'$ trials, i.e., $\Prob(\mathrm{Bin}(n',\bar p)\leq Z)=e^{-\Lambda}$, and $a\coloneqq\bar p-Z/n'$. If $a>0$, Hoeffding's inequality gives
\begin{equation}\label{eq:hoeffding}
\begin{aligned}
e^{-\Lambda}&=\Prob\big(\mathrm{Bin}(n',\bar p)-n'\bar p\leq-n'a\big)\leq\exp(-2n'a^2)\\
&\Longrightarrow\ a\leq\sqrt{\Lambda/(2n')}.
\end{aligned}
\end{equation}
The same bound holds for a lower bound $\underline p$, defined by $\Prob(\mathrm{Bin}(n',\underline p)\geq Z)=e^{-\Lambda}$, with $a\coloneqq Z/n'-\underline p$. The call uses $n'=N$ trials for $p_u$ (with $Z=N_\text{out}$) and $n'=N_{\sX_c}\geq N/2$ trials for the bounds of Theorem~\ref{thm:learning-ambiguity-ball} (with $Z=Z_i$), so \eqref{eq:hoeffding} gives deviations of at most $\sqrt{\Lambda/N}=t$. Hence $p_u\leq N_\text{out}/N+t\leq\tau/2+t$, and the bounds of Theorem~\ref{thm:learning-ambiguity-ball} satisfy $q_i-\evlowerb{i}\leq t$ and $\evupperb{i}-q_i\leq t$, with $q_i\coloneqq Z_i/N_{\sX_c}$.

Any feasible $(\omega,\gamma)$ of \eqref{eq:def-epsilon} thus has $\sum_i\gamma_{ii}\geq\sum_i\min\{\omega_i,q_i\}=1-\tfrac12\norm{\omega-q}_1\geq1-Mt/2$. By \eqref{eq:bisection}, the diagonal cost coefficients are at most $(2\kappa hM^{-1/e})^\rho$, and all others at most $\operatorname{diam}(B)^\rho\leq(\kappa h)^\rho$ (empty cells have zero coefficients, and atoms without mass play no role). As the diagonal mass is at most $1$ and the off-diagonal mass at most $Mt/2$, $\epsilon^\rho\leq(2\kappa hM^{-1/e})^\rho+(\kappa h)^\rho Mt/2$, and the claim follows from $(a^\rho+b^\rho)^{1/\rho}\leq a+b$ and $h\leq r$.
\end{proof}

\paragraph{Auxiliary result $\#3$: tight certificates.}
Finally, we show that the learner can be made arbitrarily tight along a sequence of calls. Consider stages $s \in \natNum$. In the setting of Lemma~\ref{lem:consistency}, we index by $s$ each quantity in stage $s$. Further, for a fixed $c\in(0,1)$, let $N_s = (1-c)\bar N_s$ and $n_s=c\bar N_s\in\natNum$ (so that $\bar N_s = N_s + n_s$). We assume the following \emph{schedule conditions}, as $s\to\infty$:
\[
\bar N_s, M_s \to \infty, \quad r_sM_s^{-1/e}\to0 \quad \text{and} \quad M_s^2\Lambda_sr_s^{2\rho}=o(N_s).
\]
For instance, $\bar N_s=2s$ with $N_s=n_s=s$ (i.e., $c=\frac12$), $M_s=\lceil s^{1/4}\rceil$ and $\beta_s'\geq a/s^2$ for some $a>0$ (e.g., a share of the polynomial schedule $\beta_s=6\beta/(\pi^2s^2)$) satisfy them: $s^{1/4}\leq M_s\leq2s^{1/4}$, $\Lambda_s=O(\ln s)$ and $r_s=O(\sqrt{\ln s})$, so $r_sM_s^{-1/e}=O(s^{-1/(4e)}\sqrt{\ln s})\to0$ and $M_s^2\Lambda_sr_s^{2\rho}=O(s^{1/2}(\ln s)^{1+\rho})=o(s)=o(N_s)$.

\begin{lemma}[Tight certificates]\label{lem:tight}
Under the schedule conditions above, for every $\tau\in(0,1]$ there is $s_\tau$ such that, for all $s\geq s_\tau$ and every $\sigma$-sub-Gaussian distribution on $\realNum^e$, the $s$-th call returns a radius $\epsilon\leq\tau$ and an escaped-mass upper bound $p_u\leq\tau$ with probability at least $q_s^\tau\coloneqq1-4e/(\tau (n_s+1))-1/\bar N_s$, and $q_s^\tau\to1$ as $s\to \infty$.
\end{lemma}
\begin{proof}
Fix $\tau\in(0,1]$. Since $M_s^2r_s^{2\rho}\geq(2\sigma\sqrt{\ln2})^{2\rho}$, the schedule conditions give $t_s^2=\Lambda_s/N_s\to0$, and also $r_sM_s^{-1/e}\to0$ and $r_s(M_st_s)^{1/\rho}=\big(M_s^2\Lambda_sr_s^{2\rho}/N_s\big)^{1/(2\rho)}\to0$. Hence there is $s_\tau$ such that, for all $s\geq s_\tau$, $t_s\leq\tau/2$ and $\kappa r_s\big(2M_s^{-1/e}+(M_st_s)^{1/\rho}\big)\leq\tau$. For such $s$ and any $\sigma$-sub-Gaussian distribution, Lemma~\ref{lem:consistency} gives $p_u\leq\tau/2+t_s\leq\tau$ and $\epsilon\leq\kappa r_s\big(2M_s^{-1/e}+(M_st_s/2)^{1/\rho}\big)\leq\tau$ with probability at least $1-4e/(\tau(n_s+1))-1/\bar N_s= q_s^\tau$, and $q_s^\tau\to1$ as $\bar N_s\to\infty$.
\end{proof}

In Algorithm~\ref{alg:learn-to-propagate-rrt}, the two learning calls of stage $s$ are instances of the $s$-th call above: the root call, with $(e,\bar N_s,M_s,\beta_s')=(d,N(s),M(s),\zeta\beta_s)$, and the noise call, with $(e,\bar N_s,M_s,\beta_s')=(w,N_\vomega(s),M_\vomega(s),(1-\zeta)\beta_s)$. We assume that, for both, a fraction $c\in(0,1)$ of the samples is used for pre-training and that the schedule conditions hold.

Before moving on, we need to introduce two new notions. First, by Assumption~\ref{assumption:robust-feasibility}, there is $p^\dagger>0$ such that $\sup_{\ProbQ\in\ball_\Delta(\Prob^{\tilde{\bm u}_{k-1}}_{x_k})}\ProbQ(\sO)\leq\varepsilon_\sO-p^\dagger$ for $k\in[\bar H]$ and $\sup_{\ProbQ\in\ball_\Delta(\Prob^{\tilde{\bm u}_{\bar H-1}}_{x_{\bar H}})}\ProbQ(\sX\setminus\sT)\leq\varepsilon_\sT-p^\dagger$.

Further, for $\eta>0$, define $\sU_\eta(u)\coloneqq\{u'\in\sU:\norm{u'-u}\leq\eta\}$. A controller $(u_0,\dots,u_{j-1})$ is \emph{$\eta$-aligned (of length $j$)} if $u_i\in\sU_\eta(\tilde u_i)$ for all $i<j$. Coupling the trajectories of an $\eta$-aligned controller $(u_0,\dots,u_j)$ of length $j+1\leq\bar H$ and of $\tilde{\bm u}_j$ through the same $x_0$ and noise sequence, the Lipschitz continuity of $f$ gives
\begin{equation}\label{eq:aligned}
\wasserstein_\rho\big(\Prob^{(u_0,\dots,u_j)}_{x_{j+1}},\Prob^{\tilde{\bm u}_j}_{x_{j+1}}\big)\leq C\eta
\end{equation}
for a constant $C$ depending only on $\sL_f$ and $\bar H$. Finally, let $p_\eta\coloneqq\min_{k<\bar H}\mathrm{Leb}(\sU_\eta(\tilde u_k))/\mathrm{Leb}(\sU)$, which is positive as $\sU$ is bounded and has positive Lebesgue measure in every ball centered at its points.

\paragraph{Step 4: a stage advances the forest.}
Let $\mathbb F_s$ be the forest at the beginning of stage $s$ and $\sF_s'$ the $\sigma$-algebra generated by the data consumed and the random draws made up to the end of stage $s$. Let the largest $\eta$-aligned depth in the forest $\mathbb F_s$ be $J_s\coloneqq\max\{j\leq\bar H-1:\mathbb F_s\text{ has a node whose backtrack is }\eta\text{-aligned of length }j\}$. Let $Y_s\in\sF_s'$ be the event that the algorithm returns at stage $s$ or $J_{s+1}>J_s$.

\begin{lemma}[Advance]\label{lem:advance}
Fix $\eta>0$ and $\bar\tau\in(0,1]$ such that $2\bar\tau+C\eta\leq\Delta$ and $2\bar\tau\leq p^\dagger$. On the event that the algorithm has not returned before stage $s$, under the schedule conditions, there is $s_0\geq2$ such that, for all $s\geq s_0$, $\Prob(Y_s\mid\sF_{s-1}')\geq \frac{\lambda_1\lambda_2p_\eta^K}{2(N_\text{iter}+1)s}$, where $K$ is the number of candidate controls sampled by \texttt{Propagate}.
\end{lemma}
\begin{proof}
First, for any $\tau \in (0,1]$, define $q_s$ as the sum of the probabilities $q_s^{\tau}$ of Lemma~\ref{lem:tight} for the root call and for the noise call of stage $s$, minus $1$, i.e. $q_s = q_s^{\text{root},\tau}+q_s^{\text{noise},\tau}-1$. As $q_s\to1$, $\beta_s\to0$ and $M(s)\to\infty$, there is $s_0\geq2$ such that, for all $s\geq s_0$, $s\geq s_{\tau}$ for both calls, $q_s-\beta_s\geq\frac12$ and $2\kappa \norm{\sX}_\infty M(s)^{-1/d}\leq\tau$.

Work on the event that the algorithm has not returned before stage $s$. Let $\nu^\star\in\mathbb F_s$ be a node whose backtrack $\bm u_s$ is $\eta$-aligned of length $J_s$. Let $G_s$ be the event intersecting:
(1) \texttt{SelectRoot} returns $\nu^\star$. Since $\mathbb F_s$ has $s-1$ trees with at most $N_\text{iter}+1$ nodes each, and \texttt{SelectRoot} draws uniformly a tree and a node of it with probability $\lambda_1,\lambda_2$, respectively, this has probability at least $\lambda_1\lambda_2/(s(N_\text{iter}+1))$;
(2) both learning calls of stage $s$ are valid and return a radius and an escaped-mass upper bound at most $\tau$. Given (1), $\bm u_s$ is fixed and the data of stage $s$ are fresh i.i.d.\ samples of $\Prob^{\bm u_s}_{x_{J_s}}$, which is sub-Gaussian as $\sX$ bounded, and $\Prob_\vomega$ also sub-Gaussian. So the calls are valid with probability at least $1-\beta_s$ (same step) and, by Lemma~\ref{lem:tight}, tight with probability at least $q_s$ -- thus by the union bound, they hold together with probability at least $q_s-\beta_s\geq \frac{1}{2}$;
(3) the $K$ candidates drawn by \texttt{Propagate} at the first iteration lie in $\sU_\eta(\tilde u_{J_s})$, which has probability at least $p_\eta^K$. Hence, there exists $s_0$ such that $\Prob(G_s\mid\sF_{s-1}')\geq \frac{\lambda_1\lambda_2p_\eta^K}{2(N_\text{iter}+1)s}$ for any $s\geq s_0$. We now need to show that $G_s\subseteq Y_s$ for $s \geq s_0$.

On $G_s$, by construction, $(\bm u_s,u^\star)$ is $\eta$-aligned of length $J_s+1$. It only remains to check whether this node will be accepted into the tree. From our propagation, $f_{u^\star} \# \ball_{\tau}(\ProbHat^{\bm u_s}_{x_{J_s+1}})$ has a radius of at most $\sL \tau$, where $\sL$ is some finite Lipschitz constant. Also, the propagated escaped mass is at most $2 \tau$. The compression step adds at most $2\kappa \norm{\sX}_\infty M(s)^{-1/d}$, which is at most $\tau$ from our first paragraph. In particular, if we take $\tau = \frac{\bar\tau}{2(\sL+1)}$, it holds that both the radius after compression and the escaped mass upper bound are at most $\bar\tau$.  

Then, $\wasserstein_\rho \big( \Prob^{(\bm u_s, u^\star)}_{x_{J_s+1}}, \Prob^{\bm \tilde u_{J_s}}_{x_{J_s+1}} \big) \leq C \eta \leq \Delta - 2\bar\tau$ using \eqref{eq:aligned} and the assumption on $\eta$. To conclude, from Lemma~\ref{lem:transfer}, we have that for this constructed node $\pi(\sO)\leq\varepsilon_\sO-p^\dagger+2\bar\tau\leq\varepsilon_\sO$, thus passing \texttt{ValidityCheck}. If $J_s+1<\bar H$, the new node is added to the tree, which joins the forest at the end of the stage, so $J_{s+1}\geq J_s+1>J_s$. If $J_s+1=\bar H$, the same computation with $\sS=\sX\setminus\sT$ and the second inequality defining $p^\dagger$ gives $\pi(\sX\setminus\sT)\leq\varepsilon_\sT-p^\dagger+2\bar\tau\leq\varepsilon_\sT$, so the new node also passes \texttt{TargetCheck} and the algorithm returns at stage $s$.
\end{proof}
\begin{proof}[Proof of Theorem~\ref{thm:probabilistic-completeness}]
Let $\mathsf{Ret}$ be the event that the algorithm returns. We show that $\Prob(\mathsf{Ret})=1$, i.e., that the algorithm returns within finitely many stages. Let $c_\star = \frac{\lambda_1\lambda_2p_\eta^K}{2(N_\text{iter}+1)}$.

On $\mathsf{Ret}^c$, Lemma~\ref{lem:advance} holds for all $s\geq s_0$, so $\sum_s\Prob(Y_s\mid\sF_{s-1}')\geq\sum_{s\geq s_0}c_\star/s=\infty$. By L\'evy's extension of the Borel--Cantelli lemma (for events $Y_s\in\sF_s'$, almost surely $\sum_s\Prob(Y_s\mid\sF_{s-1}')=\infty$ if and only if $Y_s$ occurs infinitely often), $Y_s$ then occurs infinitely often, almost surely on $\mathsf{Ret}^c$. But without returning, $Y_s$ means $J_{s+1}>J_s$, which can happen at most $\bar H-1$ times, as $J_s\in\{0,\dots,\bar H-1\}$ is nondecreasing. Hence $\Prob(\mathsf{Ret}^c)=0$.
\end{proof}

\section{Promising controller selection heuristics}\label{section:appendix-heuristics}

In this Section, we present the heuristics employed by \texttt{SelectRoot} (Algorithm~\ref{alg:select-root}) to select a promising sequence of controls after each learning stage. This heuristic is based on a simple trajectory tracking rule; however, as highlighted in the main text, this represents only one of many possible strategies. Intuitively, we first execute a standard RRT on a deterministic variant of System~\ref{eq:system_dynamics} to rapidly generate multiple reference trajectories to the target set -- a step we term \emph{Informed Search}. We then select one of these trajectories and attempt to track using Algorithm~\ref{alg:learn-to-propagate-rrt}.

\subsection{Informed Search Heuristic}
\begin{figure*}[t]
    \centering
    \begin{subfigure}{0.4\textwidth} 
        \centering
        \includegraphics[width=\textwidth]{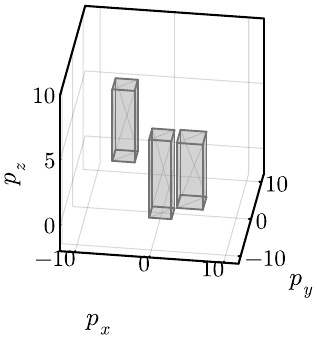}
        \caption{Tree}
        \label{fig:di_image1}
    \end{subfigure}
    \begin{subfigure}{0.4\textwidth} 
        \centering
        \includegraphics[width=\textwidth]{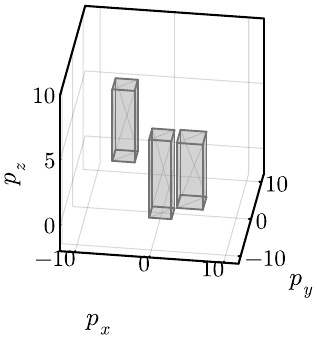}
        \caption{Reference trajectory}
        \label{fig:di_image2}
    \end{subfigure}
    \caption{\emph{Informed search} heuristic for \emph{Double Integrator} (only the three first state dimensions are depicted).}
    \label{fig:double_integrator-is}
\end{figure*}

Prior to executing Algorithm \ref{alg:learn-to-propagate-rrt}, we run a standard RRT on a deterministic version of System~\ref{eq:system_dynamics}, wherein the initial state is set to the empirical mean of $\Prob_{x_0}$ (estimated via sampling) and the process noise to zero (the mean of $\Prob_\vomega$ in all our benchmarks). The determinism reduces the computational overhead to a negligible fraction of that incurred by the full stochastic algorithm. Additionally, we utilize goal bias (probability $0.3$) in the RRT, as we observe empirically that it significantly improves its efficiency, and inflate its unsafe boxes by a margin (Table~\ref{tab:hyperparameters}) to leave clearance around obstacles. Figure \ref{fig:double_integrator-is} illustrates the tree generated by \emph{informed search} after $20{,}000$ iterations for the \emph{Double Integrator} environment.

From the set of all trajectories discovered during the \emph{Informed Search} phase, we then select a reference trajectory that maximizes the minimum distance to the avoid sets across all time steps. The resulting selected trajectory is also depicted in Figure \ref{fig:double_integrator-is} for the \emph{Double Integrator}. This reference trajectory is computed once per benchmark (with a fixed seed) and shared across all trials and methods. Its computation time is excluded from the reported planning times.

\subsection{\texttt{SelectRoot}: Promising Controller and Node Selection}\label{subsection:select-promising-trajectory}
The promising node of a tree, returned by \texttt{SelectRoot} (Algorithm~\ref{alg:select-root}) when it does not explore, is a non-root node $n^\star$ of the tree. The stage then re-learns its root from the state distribution induced by the controller $\texttt{backtrack}(n^\star)$ leading to $n^\star$ (Lines~\ref{step:select-root}--\ref{step:learn-root}). Let $\bar x_n$ be the mean of the center distribution of a node $n$, $\bar x_\text{root}$ that of the root, and $\bm{x}_\text{ref} = (\tilde{x}_1,\dots,\tilde{x}_T)$ the reference trajectory from \emph{Informed Search}. The rule has two cases:
\begin{enumerate}
    \item[(a)] \emph{Target nearly certified.} If some node of the tree has $1-\pi(\sX\setminus\sT)>0.5$, i.e., its \texttt{TargetCheck} bound (Eq.~\eqref{eq:node-bound}) certifies a target probability above $0.5$, select the non-root node whose center is closest to the reach sets, measured by the probability-weighted average distance from its support points to the nearest reach-set center.
    \item[(b)] \emph{Reference tracking.} Otherwise, select the node that has advanced the farthest along the reference trajectory $\bm{x}_\text{ref}$ while staying close to it. Only the part of $\bm{x}_\text{ref}$ ahead of the root matters, i.e., the states after the one closest to $\bar x_\text{root}$, and the selection has two steps:
    \begin{itemize}
        \item \emph{Candidates.} Keep the $50$ nodes whose means are the farthest from $\bar x_\text{root}$, which favors exploration.
        \item \emph{Progress.} A candidate $n$ \emph{reaches} the state $\tilde x_i$ if $\norm{\bar x_n - \tilde x_i} \leq 3\bar d$, where $\bar d$ is the average distance between consecutive states ahead. Select the candidate that reaches the state with the largest index $i$, i.e., the farthest along $\bm{x}_\text{ref}$.
    \end{itemize}
\end{enumerate}
The heuristic balances exploration (nodes far from the root) and guidance (progress along the reference), and has demonstrated strong empirical performance across the experiments considered in this work.

\section{Experimental details}\label{section:appendix-experiments-details}

We summarize the system dynamics, and environment specifications used in the experiments presented in Section \ref{section:experimental-results}.

\begin{table*}[t]
\centering
\small
\setlength{\tabcolsep}{6pt}
\renewcommand{\arraystretch}{1.15}
\begin{tabular}{lccccc}
\toprule
 & \emph{Switched System} & \emph{AV Maze} & \emph{Unicycle Maze} & \emph{Tractor Trailer} & \emph{Double Integrator} \\
\midrule
Pre-training samples $N_\text{pre}^{(1)}$ & $2{,}000$ & $2{,}000$ & $2{,}000$ & $1{,}000$ & $1{,}000$ \\
Training samples $N^{(1)}$ & $2{,}000$ & $20{,}000$ & $10{,}000$ & $10{,}000$ & $10{,}000$ \\
Steering candidates & $100$ & $100$ & $100$ & $500$ & $1{,}000$ \\
Informed search iterations & $20{,}000$ & $20{,}000$ & $20{,}000$ & $10{,}000$ & $20{,}000$ \\
Unsafe-box inflation & $0$ & $0.1$ & $0.05$ & $1.0$ & $1.0$ \\
\bottomrule
\end{tabular}
\caption{Hyperparameters per benchmark.}
\label{tab:hyperparameters}
\end{table*}

\subsection{Autonomous Vehicle Maze \cite{abate2025arch}}\label{section:appendix-av-maze}

The \emph{AV Maze} benchmark from \cite{abate2025arch} uses a reduced autonomous vehicle model with state
\[
x_k = (p_{x, k}, p_{y, k}, \theta_k) \in [0,4] \times [0,10] \times [-\pi-0.4,\pi+0.4],
\]
where $(p_{x, k},p_{y, k})$ is position and $\theta_k$ is the heading angle, both at time $k$. The control input is given by $u_k = (v_k,\delta_k) \in [-1,1]^2$, where $v_k$ represents the longitudinal velocity and $\delta_k$ the steering angle. Let
\[
\alpha_k = \arctan\!\left(\frac{\tan(\delta_k)}{2}\right).
\]
We consider the dynamics to be described by an additive noise dynamical process given by $x_{k+1}=f(x_k, u_k, \omega_k)= g(x_k, u_k) + \vomega_k$, where the function $g$ is given by
\[
p_{x,k+1} = p_{x,k} + \tau v_k \frac{\cos(\theta_k + \alpha_k)}{\cos(\alpha_k)},
\]
\[
p_{y,k+1} = p_{y,k} + \tau v_k \frac{\sin(\theta_k + \alpha_k)}{\cos(\alpha_k)},
\]
\[
\theta_{k+1} = \operatorname{wrap}_{(-\pi,\pi]}\!\left(\theta_k + \tau v_k \tan(\delta_k)\right),
\]
with sampling time $\tau = 0.5$. For any $u\in \sU$, the Lipschitz constant of $g(\cdot,u)$ is
\[
\sL_{g_u} = \sqrt{1+\frac{\sigma^2}{2}+\frac{\sigma}{2}\sqrt{\sigma^2+4}},
\]
\[
\sigma = \frac{0.5\,|v|}{\cos(\alpha)} = 0.5\,|v|\sqrt{1+\frac{\tan^2(\delta)}{4}},
\]
with $\alpha$ being the slip angle defined above. This constant is tight: the Jacobian of $g(\cdot,u)$ is $I + w e_3^\top$ with $\|w\|_2 = \sigma$ independently of the state, and the expression above is its spectral norm. Over $\sU$, it ranges between $[1, 1.3659]$.

The unsafe set is defined as $\sO = \sO_1 \cup  \sO_2 \cup \sO_3$, where
\[
\sO_1 = [1.00,1.12] \times [0,8.6] \times [-\pi-0.4,\pi+0.4],
\]
\[
\sO_2 =[2.24,2.36] \times [0,4.6] \times [-\pi-0.4,\pi+0.4],
\]
\[
\sO_3 =[2.24,2.36] \times [6.4,10.0] \times [-\pi-0.4,\pi+0.4].
\]
The reach set is given by $\mathcal{T} = [2.85,3.75] \times [0.22,0.97] \times [-\pi-0.4,\pi+0.4]$. Risk thresholds are $\varepsilon_{\sO} = 0.05$ and $\varepsilon_{\sT} = 0.05$. The unknown distributions $\Prob_{x_0}, \Prob_\vomega$ are respectively set to a Dirac distribution at $(0.4,0.4,0)$, and a zero-mean Gaussian with a diagonal covariance matrix defined as $\text{diag}(10^{-10},10^{-10},0)$, i.e., the heading angle $\theta$ is unaffected by process noise.

\subsection{Unicycle Maze}\label{section:appendix-unicycle-maze}

The \emph{Unicycle Maze} benchmark reuses the reduced kinematic-vehicle dynamics of Appendix~\ref{section:appendix-av-maze} (same $g$, same $\tau = 0.5$, same Lipschitz-constant expression) over a larger, more cluttered $9.8\times9.8$ arena with state
\[
x_k = (p_{x, k}, p_{y, k}, \theta_k) \in [-4.9,4.9] \times [-4.9,4.9] \times [-\pi-0.4,\pi+0.4],
\]
and control $u_k = (v_k,\delta_k) \in [-1,1]^2$. The unsafe set is defined as $\sO = \sO_1 \cup \sO_2 \cup \sO_3 \cup \sO_4 \cup \sO_5$, where
\[
\sO_1 = [1.5,2.0] \times [1.0,4.2] \times [-\pi-0.4,\pi+0.4],
\]
\[
\sO_2 = [-3.0,2.0] \times [1.0,1.5] \times [-\pi-0.4,\pi+0.4],
\]
\[
\sO_3 = [-3.0,2.0] \times [-1.0,-0.5] \times [-\pi-0.4,\pi+0.4],
\]
\[
\sO_4 = [1.5,2.0] \times [-3.5,-2.1] \times [-\pi-0.4,\pi+0.4],
\]
\[
\sO_5 = [-4.3,2.0] \times [-3.5,-3.0] \times [-\pi-0.4,\pi+0.4],
\]
forming a corridor-and-wall maze. The reach set is given by $\sT = [-1.1,1.0] \times [-2.6,-1.5] \times [-\pi-0.4,\pi+0.4]$. Risk thresholds are $\varepsilon_{\sO} = \varepsilon_{\sT} = 0.05$. The unknown distributions $\Prob_{x_0}, \Prob_\vomega$ are respectively set to a Dirac distribution at $(4.0,4.0,0)$, and a zero-mean Gaussian with a diagonal covariance matrix defined as $\text{diag}(5\times10^{-7},5\times10^{-7},5\times10^{-7})$.

\subsection{3D Double Integrator}

The \emph{3D Double Integrator} benchmark consists of a six-dimensional model with state described by
\begin{multline*}
x_k = (p_{x, k},p_{y, k},p_{z, k},v_{x, k}, v_{y, k},v_{z, k}) \\ \in [-12,12] \times [-12,12] \times [-2,10] \times [-4,4] \times [-4,4] \times [-3,3],
\end{multline*}
representing $3$D positions and velocities, and control $u_k = (a_{x, k},a_{y, k},a_{z, k}) \in [-1,1]^3$. We consider the dynamics to be described by an additive noise dynamical process given by $x_{k+1}=f(x_k, u_k, \omega_k)= g(x_k, u_k) + \vomega_k = A x_k + B u_k + \vomega_k$, where
\[
A =
\begin{bmatrix}
1 & 0 & 0 & \tau & 0 & 0 \\
0 & 1 & 0 & 0 & \tau & 0 \\
0 & 0 & 1 & 0 & 0 & \tau \\
0 & 0 & 0 & 1 & 0 & 0 \\
0 & 0 & 0 & 0 & 1 & 0 \\
0 & 0 & 0 & 0 & 0 & 1
\end{bmatrix}, \qquad
B =
\begin{bmatrix}
\frac{1}{2}\tau^2 & 0 & 0 \\
0 & \frac{1}{2}\tau^2 & 0 \\
0 & 0 & \frac{1}{2}\tau^2 \\
\tau & 0 & 0 \\
0 & \tau & 0 \\
0 & 0 & \tau
\end{bmatrix},
\]
with sampling time $\tau=0.3$. For any $u\in \sU$, the Lipschitz constant of $g(\cdot,u)$ is the spectral norm of the state-transition matrix $\sL_{g_u} = \|A\| \approx 1.1612$. The unsafe set is defined as $\sO = \sO_1 \cup \sO_2 \cup \sO_3$, where
\[
\sO_1 = [-1.5,1.5] \times [-1.5,1.5] \times [-2.0,4.0] \times [-4,4] \times [-4,4] \times [-3,3],
\]
\[
\sO_2 = [-7.0,-4.0] \times [2.5,5.5] \times [1.0,6.5] \times [-4,4] \times [-4,4] \times [-3,3],
\]
\[
\sO_3 = [3.0,6.5] \times [-7.0,-3.5] \times [0.5,5.5] \times [-4,4] \times [-4,4] \times [-3,3].
\]
The reach set is given by $\sT = [6.5,9.5] \times [6.5,9.5] \times [5.0,8.0] \times [-1.5,1.5]^3$. Risk thresholds are $\varepsilon_{\mathcal{O}} = \varepsilon_{\mathcal{T}} = 0.05$. The unknown distributions $\Prob_{x_0}, \Prob_\vomega$ are respectively set to a Dirac distribution at $(-8.5,-8.5,0,0,0,0)$, and a zero-mean Gaussian with a diagonal covariance matrix defined as $\text{diag}(0,0,0,10^{-8},10^{-8},10^{-8})$.

\subsection{Switched Systems (5 Modes) \cite{gracia2025efficient}}

The \emph{Switched Systems (5 Modes)} benchmark from \cite{gracia2025efficient} considers a two-dimensional state
\[
x_k = (p_{x,k}, p_{y,k}) \in [-2,2]^2,
\]
with scalar control input $u_k \in [0,5]$. The control selects one of five linear modes via
\[
\sigma(u_k) =
\begin{cases}
1, & 0 \le u_k < 1,\\
2, & 1 \le u_k < 2,\\
3, & 2 \le u_k < 3,\\
4, & 3 \le u_k < 4,\\
5, & 4 \le u_k \le 5.
\end{cases}
\]
We consider the dynamics to be described by an additive noise dynamical process given by
\[
x_{k+1} = f(x_k,u_k,\omega_k) = g(x_k,u_k) + \vomega_k = A_{\sigma(u_k)} x_k + \vomega_k,
\]
where the mode-dependent matrices are
\[
A_1 =
\begin{bmatrix}
0.79 & 0.035 \\
0.0 & 0.825
\end{bmatrix},
\qquad
A_2 =
\begin{bmatrix}
0.79 & 0.175 \\
0.0 & 0.825
\end{bmatrix},
\]
\[
A_3 =
\begin{bmatrix}
0.79 & 0.0 \\
0.175 & 0.825
\end{bmatrix},
\]
\[
A_4 =
\begin{bmatrix}
1.0 & 0.2 \\
-0.2 & 1.0
\end{bmatrix},
\qquad
A_5 =
\begin{bmatrix}
1.0 & -0.2 \\
0.2 & 1.0
\end{bmatrix}.
\]
For any $u \in \sU$, the Lipschitz constant of $g(\cdot,u)$ is the spectral norm (largest singular value) of the active mode matrix, i.e.\ $\sL_{g_u} \approx 0.8324, 0.9015, 0.9015, 1.0198, 1.0198$ for modes $1$ to $5$ respectively. 

The unsafe set is defined as $\sO = \sO_1 \cup \sO_2 \cup \sO_3 \cup \sO_4$, where
\[
\sO_1 = [-2.0,-1.0] \times [-2.0,-1.0], \qquad
\sO_2 = [-1.0,0.0] \times [0.0,1.0],
\]
\[
\sO_3 = [0.0,1.0] \times [-1.0,0.0], \qquad
\sO_4 = [1.0,2.0] \times [1.0,2.0],
\]
together with the complement of the support $[-2,2]^2$. The reach set is given by
\[
\sT_1 = [0.0,1.0] \times [0.0,1.0], \qquad
\sT_2 = [-1.0,0.0] \times [-1.0,0.0].
\]
Risk thresholds are $\varepsilon_{\sO} = \varepsilon_{\sT} = 0.05$. The unknown distributions $\Prob_{x_0}, \Prob_\vomega$ are respectively set to a Dirac distribution at $(-1.5,1.0)$, and a zero-mean Gaussian with a diagonal covariance matrix defined as $\operatorname{diag}(10^{-6},10^{-6})$.

\subsection{Tractor-Trailer NN \cite{bos2023mpc}}\label{section:appendix-tractor-trailer}

The \emph{Tractor-Trailer NN} benchmark uses a four-dimensional state
\begin{multline*}
x_k = (p_{x,k}, p_{y,k}, \theta_k, \alpha_k) \\ \in [-4.0,10.0] \times [-4.0,10.0] \times [-3.15,3.15] \times [-1.57,1.57],
\end{multline*}
where $(p_{x,k},p_{y,k})$ is the position of the tractor, $\theta_k$ its heading, and $\alpha_k$ the hitch angle between the tractor and the trailer, with control $u_k = (v_k,\omega_k) \in [-2.0,2.0] \times [-1.0,1.0]$ representing the longitudinal velocity and the steering rate. The nominal dynamics are given by a residual neural network,
\[
x_{k+1} = f(x_k,u_k,\omega_k) = g(x_k,u_k) + \vomega_k = x_k + \mathcal{NN}(\theta_k,\alpha_k,v_k,\omega_k) + \vomega_k,
\]
where $\mathcal{NN}$ is a fully connected network with architecture $4$-$64$-$64$-$4$, $\tanh$ activations on the two hidden layers, and a linear output layer, trained offline to approximate the discrete-time residual kinematics of the truck-trailer model of \cite{bos2023mpc}; only the heading and hitch angle $(\theta_k,\alpha_k)$ and the control $(v_k,\omega_k)$ are fed to the network, while the position increments $(\Delta p_{x,k},\Delta p_{y,k})$ are read off its output. The Lipschitz constant of $g(\cdot,u)$ used by our implementation is bounded by
\[
\sL^{\text{naive}}_{g_u} = 1 + \|W_3\|_2\, \|W_2\|_2\, \|W_1^{(\theta,\alpha)}\|_2 \approx 1.6468,
\]
where $W_1$, $W_2$, and $W_3$ are the weight matrices of the three linear layers and $W_1^{(\theta,\alpha)}$ denotes the two columns of $W_1$ acting on $(\theta_k,\alpha_k)$ (the position channels $(p_{x,k},p_{y,k})$ do not feed $\mathcal{NN}$, and $(v_k,\omega_k)$ are controls rather than state variables). The leading $1$ accounts for the residual (skip) connection: the Jacobian of $g(\cdot,u)$ is $I_4 + M$, where $M$ is supported on the $(\theta,\alpha)$ columns and satisfies $\|M\|_2 \le \|W_3\|_2 \|W_2\|_2 \|W_1^{(\theta,\alpha)}\|_2$ since $\tanh$ is $1$-Lipschitz. In particular $\sL_{g_u} \ge 1$, as perturbing only $(p_{x,k},p_{y,k})$ leaves the network input unchanged and is therefore propagated isometrically.

This bound is conservative -- it assumes all $128$ hidden units saturate simultaneously -- and, compounded once per time step over the ${\approx}90$ steps of this benchmark, it renders the propagated ambiguity sets vacuous. We therefore certify the constant with the LipSDP semidefinite relaxation of \cite{fazlyab2019efficient}, applied to the residual map $g(\cdot,u)$ (see \texttt{src/lipschitz.jl}). This yields $\sL^{\text{LipSDP}}_{g_u} \approx 1.2096$, against ${\approx}1.1046$ for the numerically estimated true constant. All \emph{Tractor-Trailer NN} results use $\sL^{\text{LipSDP}}_{g_u}$; with $\sL^{\text{naive}}_{g_u}$ the success rate collapses to $5\%$.

The unsafe set is defined as $\sO = \sO_1$, where
\[
\sO_1 = [2.0,3.0] \times [2.0,3.0] \times [-3.15,3.15] \times [-1.57,1.57],
\]
together with the complement of the support. The reach set is given by
\[
\sT = [4.0,7.0] \times [4.0,7.0] \times [-3.15,3.15] \times [-1.57,1.57].
\]
Risk thresholds are $\varepsilon_{\sO} = \varepsilon_{\sT} = 0.05$. The unknown distributions $\Prob_{x_0}, \Prob_\vomega$ are respectively set to a Dirac distribution at $(-0.5,1.5,0.78,0.0)$, and a zero-mean Gaussian with a diagonal covariance matrix defined as $\operatorname{diag}(10^{-8},10^{-8},10^{-8},10^{-8})$.

\subsection{Hyperparameters}\label{section:appendix-hyperparameters}
Table~\ref{tab:hyperparameters} lists the hyperparameters of Algorithm~\ref{alg:learn-to-propagate-rrt} that are not given in Section~\ref{section:experimental-results}. At stage $s$, both calls to \texttt{LearnAmbiguitySet} (root and noise) use $s\,N_\text{pre}^{(1)}$ pre-training and $s\,N^{(1)}$ training samples, i.e., $N(s)=N_\vomega(s)=s\,(N_\text{pre}^{(1)}+N^{(1)})$. Line~\ref{step:before-steer} of Algorithm~\ref{alg:propagate-node} takes the best of the number $K$ of uniformly sampled steering candidates indicated in the table. The last two rows are the settings of the informed search of Appendix~\ref{section:appendix-heuristics}.

% \subsection{CartPole NN}

% The \emph{CartPole NN} benchmark uses a four-dimensional state
% \[
% x_k = (x_k,\dot{x}_k,\theta_k,\dot{\theta}_k) \in [-2.4,2.4] \times [-10,10] \times [-1.6,1.6] \times [-8,8],
% \]
% with scalar control input $u_k \in [-25,25]$. The nominal dynamics are represented by a residual neural network, so that
% \[
% x_{k+1} = f(x_k,u_k,\omega_k) = g(x_k,u_k) + \vomega_k = x_k + \mathcal{NN}(x_k,u_k) + \vomega_k,
% \]
% where $\mathcal{NN}$ is a fully connected neural network with architecture $5$-$64$--$64$-$4$ and $\tanh$ activations in the hidden layers, that was trained to learn the residual dynamics of a classical cartpole (see e.g. \cite{Sutton1998}). For any $u \in \sU$, the Lipschitz constant of $g(\cdot,u)$ is bounded by
% \[
% \sL_{g_u} \le 1 + \|W_3\|_2 \, \|W_2\|_2 \, \|W_1\|_2 \approx 13.03,
% \]
% where $W_1$, $W_2$, and $W_3$ are the weight matrices of the three linear layers in the neural network. There is no unsafe set, and the reach set is given by
% \[
% \sT = [-0.45,0.45] \times [-10,10] \times [-0.1,0.1] \times [-8,8].
% \]
% Risk thresholds are $\varepsilon_{\sO} = \varepsilon_{\sT} = 0.05$. The unknown distributions $\Prob_{x_0}, \Prob_\vomega$ are respectively set to a Dirac distribution at $(-2.0,0.0,1.5,0.0)$, and a zero-mean Gaussian with a diagonal covariance matrix defined as $\operatorname{diag}(10^{-5},10^{-5},10^{-5},10^{-5})$.

\section{$\epsilon$-RandUP-RRT baseline}\label{section:appendix-randup-rrt}
We implement $\epsilon$-RandUP-RRT following the reference code released by \cite{wu2022robust} at \url{https://github.com/StanfordASL/randUP_RRT}, which combines the particle-based, $\epsilon$-padded convex-hull reachability check of \cite{lew2022randup} with an RRT planner: each tree node holds $N$ particles propagated under independently sampled noise trajectories, and a node (or edge) is accepted only if the convex hull of its particles avoids every padded obstacle and, to certify the target, is contained in the padded target set, both checked via a linear program. Two differences from the released code: (i) we bias sampling toward a nominal reference trajectory, computed once up front with our own risk-agnostic tree-expansion routine, rather than sampling uniformly, to aid convergence in the more cluttered mazes, and (ii) we impose a $2$-minute wall-clock budget per run, recording unresolved runs as timeouts. Following \cite{wu2022robust}, we use a padding of $\epsilon=0.3$ from their quadrotor experiment, and set $N=1{,}000$ particles for every benchmark.

\end{document}